\pdfoutput=1
\documentclass[11pt]{article}

\usepackage[preprint]{acl}

\usepackage{times}
\usepackage{latexsym}
\usepackage{acronym}
\usepackage{tabularx}
\usepackage{multirow}
\usepackage{amsmath, amsthm}

\theoremstyle{definition}
\newtheorem{definition}{Definition}[section] % [section] resets numbering per section

\usepackage[T1]{fontenc}
\usepackage[utf8]{inputenc}

\usepackage{microtype}

\usepackage{inconsolata}

\usepackage{booktabs}
\usepackage{comment}
\usepackage{bbm}
\usepackage{amssymb}
\usepackage{dsfont}
\usepackage{stfloats}
\usepackage{graphicx}
\usepackage{algorithm}
\usepackage{algpseudocode}

\usepackage[table]{xcolor}
\usepackage{xcolor}
\definecolor{ours}{RGB}{224,224,255}
\newcommand{\add}[1]{\textcolor{black}{#1}}

\definecolor{ggreen}{RGB}{16,144,64}
\newcommand{\dec}[1]{\textcolor{ggreen}{(-#1)}}

\newcommand{\dep}[1]{(-#1\%)}
\newcommand{\inp}[1]{(+#1\%)}
\acrodef{AL}[AL]{Active Learning}
\acrodef{BiLSTM}[LSTM]{Bidirectional Long Short-Term Memory}
\acrodef{CNN}[CNN]{Convolutional Neural Network}
\acrodef{CrI}[CrI]{Credible Interval}
\acrodef{DDS}[DDS]{Denoised Distant Dataset}
\acrodef{DOREMI}[DOREMI]{\textbf{DO}cument-level \textbf{R}elation \textbf{E}xtraction opti\textbf{M}izing the long ta\textbf{I}l}
\acrodef{DDS}[DDS]{Denoised Distantly Supervised Dataset}
\acrodef{DS}[DS]{Distant Supervision}
\acrodef{DocRE}[DocRE]{Document-Level Relation Extraction}
\acrodef{EM}[EM]{Expectation Maximization}
\acrodef{ER}[ER]{Evidence Retrieval}
\acrodef{GCN}[GCN]{Graph Convolutional Network}
\acrodef{GNN}[GNN]{Graph Neural Network}
\acrodef{GAT}[GAT]{Graph Attention Network}
\acrodef{HPD}[HPD]{High Posterior Density}
\acrodef{IE}[IE]{Information Extraction}
\acrodef{KBC}[KBC]{Knowledge Base Construction}
\acrodef{KG}[KG]{Knowledge Graph}
\acrodef{LLM}[LLM]{Large Language Model}
\acrodef{LSTM}[LSTM]{Long Short-Term Memory}
\acrodef{NER}[NER]{Named Entity Recognition}
\acrodef{NER+L}[NER+L]{Named Entity Recognition and Linking}
\acrodef{NLP}[NLP]{Natural Language Processing}
\acrodef{OBDM}[OBDM]{Ontology-Based Data Management}
\acrodef{PPD}[PPD]{Positive Probability Disagreement}
\acrodef{PPM}[PPM]{Positive Probability Mean}
\acrodef{RE}[RE]{Relation Extraction}
\acrodef{SC}[SC]{Selection Criteria}
\acrodef{SRS}[SRS]{Simple Random Sampling}
\acrodef{OWL}[OWL]{Web Ontology Language}
\acrodef{RDF}[RDF]{Resource Description Framework}
\title{Ontology-Driven Structural Regularization for \\ Document-Level Relation Extraction}
\author{
   \textbf{Laura Menotti\textsuperscript{1}},
   \textbf{Stefano Marchesin\textsuperscript{1}},
   \and
   \textbf{Gianmaria Silvello\textsuperscript{1}}
\\
\textsuperscript{1}Department of Information Engineering, University of Padua, Italy
\\
\texttt{\{name.surname\}@unipd.it}
}
\begin{document}
\maketitle
\begin{abstract}

\ac{DocRE} relies heavily on costly manually annotated datasets, while large distant supervision resources such as \emph{DocRED distant} remain underexploited due to noise. We show that a critical yet overlooked source of noise lies in structural inconsistencies within relational triples, including violations of ontology constraints and logical contradictions.

We introduce an ontology-driven framework to quantify and enforce structural consistency in DocRE datasets. Our analysis reveals substantial structural noise in DocRED distant and demonstrates that such inconsistencies propagate to model predictions. Enforcing structural well-formedness during training significantly reduces logical contradictions and consistently improves generalization performance.
These findings establish structural consistency as a missing axis of supervision in DocRE and highlight structural regularization as an effective strategy for leveraging distant data at scale.
\end{abstract}

\section{Introduction}

\acf{DocRE} aims to identify all semantic relations holding between entity pairs within a full document. By extracting relational triples, DocRE enables the construction of \acp{KG}, which provide structured, machine-readable knowledge to support downstream tasks such as information retrieval, question answering, data mining, and recommendation systems \cite{dong_etal-2023}. Building accurate KGs therefore depends on reliable DocRE models trained on high-quality data.

The reference benchmark for general-domain DocRE is DocRED \cite{yao_etal-2019}, constructed from Wikipedia articles. DocRED includes both manually annotated datasets and a large distantly supervised dataset. The manually curated split follows a recommend-revise paradigm and contains 5,053 documents, divided into 3,053 for training and 1,000 each for development and testing. In contrast, the DocRED distant dataset comprises 101,873 documents, annotated via \ac{DS} by aligning Wikidata triples with Wikipedia text and leveraging BERT-based named entity annotations.

%\ac{DocRE} models can be categorized into graph-based models and sequence-based models. The former construct a graph for each document where nodes are its entities and the edges represent dependencies between them~\cite{zeng_etal-2020,jain_etal-2024}. Such models usually leverage only the manual training dataset as they cannot effectively scale to the size of distantly supervised datasets. Sequence-based models consider documents as a sequence of tokens and learn contextual representation for all entity pairs. Most effective models exploit transformer-based encoders, e.g. BERT, to learn contextual embeddings~\cite{zhou_etal-2021,ma_etal-2023}. 

While distant supervision enables large-scale data construction \cite{mintz_etal-2009}, it relies on the strong assumption that if a \ac{KG} asserts a relation between two entities, then every document mentioning the pair expresses that relation. This assumption adds substantial noise. In DocRE, the issue is worse because entities co-occurring in a document are less likely to express a true relation than those in the same sentence. As a result, most state-of-the-art models rely mainly on manually annotated data or use distant data only for pre-training or distillation \cite{tan_etal-2022a}.

Recent analyses have further revealed annotation issues even in the manually curated splits. In particular, around 65\% of positive relations were missing in the original DocRED annotations \cite{huang_etal-2022}, leading to the release of ReDocRED, a revised version of the training and development datasets \cite{tan_etal-2022b}. Although existing denoising approaches attempt to mitigate DS noise through uncertainty estimation \cite{sun_etal-2023} or active learning for long-tail relations \cite{menotti_etal-2026-DOREMI}, they focus primarily on semantic correctness and model-driven pseudo-label refinement.

In contrast to model-centric denoising strategies, we adopt a \emph{data-centric} approach, focusing on improving the structural quality of training data itself. We argue that beyond semantic noise, DocRE datasets suffer from \emph{structural inconsistencies} in the extracted triples, such as violations of ontology constraints, missing inverse relations, asymmetric contradictions, and cardinality conflicts.\footnote{We use the term structural consistency to denote the well-formedness of RDF triples under ontology-imposed constraints, rather than full ontology satisfiability.} These inconsistencies represent a previously overlooked source of supervision noise that might propagate during model training.

Concretely, we operationalize structural consistency by leveraging the formal semantics of the \ac{OWL} to define explicit structural constraints over entity types and relations. Our methodology evaluates the \emph{well-formedness} of relational triples under ontology axioms and quantifies structural violations in both DocRED and ReDocRED, with particular emphasis on the DocRED distant dataset. 

We show that structural noise is substantially more prevalent in distantly supervised data than in manually curated datasets. In particular, compared to the ReDocRED manual training split, the DocRED distant dataset contains three times more invalid triples, nearly five times more missing inverse relations, and twice as many asymmetric inconsistencies. These findings suggest that structural inconsistencies constitute a systematic source of noise that can propagate during training, impairing a model’s ability to learn and generate well-formed triples. To mitigate this effect, we introduce a lightweight, model-agnostic \emph{pre-processing pipeline} that enforces structural consistency prior to training. The pipeline cleans and augments the dataset without altering model architectures, loss functions, or inference procedures, ensuring seamless integration with existing DocRE systems -- whether sequence-based or graph-based.

We show that enforcing ontological consistency acts as a form of structural regularization, improving both prediction coherence and generalization. %Empirical results confirm our hypothesis: removing structural inconsistencies not only substantially reduces logical contradictions in model predictions but also leads to consistent improvements in downstream performance. 
Our main contributions are: (1) We propose an ontology-driven framework that uses \ac{OWL} constraints to rigorously diagnose ontology-level structural consistency in DocRE datasets; (2) We deliver the first in-depth quantitative study of structural inconsistencies in DocRED distant and its denoised variants, uncovering extensive, previously unreported violations; (3) We show that enforcing structural well-formedness as a pre-processing step markedly reduces logical contradictions in model predictions and consistently boosts generalization performance across state-of-the-art DocRE models \cite{zhou_etal-2021, ma_etal-2023}.

We release the complete analysis and cleaning framework as open-source.~\footnote{\url{https://github.com/mntlra/DocRE-StructuralConsistency}} The toolkit is compatible with any dataset formatted in the DocRED schema and includes the ontology-based rules we developed, while allowing for the definition of other custom constraints.

The rest of this paper is organized as follows. Section~\ref{sec:related} reviews prior work on distant denoising and constrained \ac{RE}. Section~\ref{sec:rules} defines the structural rules used in our analysis. Section~\ref{sec:data-quality} assesses the structural consistency of DocRED and ReDocRED and examines the main causes of invalid triples. Section~\ref{sec:training} studies how structural noise affects \ac{DocRE} models during training. Finally, Section~\ref{sec:conclusion} concludes the paper.
%Section~\ref{sec:related} presents previous efforts in \ac{DocRE}, with special attention to label denoising and sequence-based models. Section~\ref{sec:methodology} describes \ac{DOREMI}.\footnote{Code and data are available as supplementary material and will be publicly released upon acceptance.} Section~\ref{sec:results} reports the performance of \ac{DOREMI} compared to other denoising strategies and an ablation study about disagreement-based sampling. To conclude, section~\ref{sec:conclusion} draws some final remarks.

\section{Related Work}
\label{sec:related}
% \lm{Distant Datasets}
% DocRED~\cite{yao_etal-2019}, UGDRE~\cite{sun_etal-2023}, and DOREMI~\cite{menotti_etal-2026-DOREMI}.
%\lm{Denoising techinques. Show that no prior work addresses syntactic correctness in DocRE datasets.}

\acf{DS} enables large-scale relation extraction by aligning entity pairs in text with triples from a reference \ac{KG}~\cite{mintz_etal-2009}. While efficient, this paradigm introduces noisy supervision due to the heuristic alignment between text and \ac{KG} facts. 

To address this issue, several works have proposed denoising strategies for \ac{DocRE}. \citet{xiao_etal-2020} introduced a multi-task framework incorporating mention-entity matching and fact alignment. ~\citet{sun_etal-2023} proposed uncertainty-guided label denoising using Monte Carlo dropout to filter low-confidence pseudo-labels. ~\citet{menotti_etal-2026-DOREMI} developed an active learning pipeline tailored to improve long-tail relation predictions. 

These approaches focus primarily on improving the semantic reliability of distant labels through model-driven refinement. In contrast, our work targets a complementary and largely unexplored dimension: structural consistency of relational triples under ontology-imposed constraints.

%\citet{xiao_etal-2020} exploits a denoising technique based on multi-task learning with mention-entity matching and fact alignment. \citet{sun_etal-2023} proposes Monte-Carlo dropout to estimate pseudo-label certainty of pre-trained model predictions and filters out low-confidence labels. This approach generates two denoised distant datasets, one created by pre-training the \ac{DocRE} model on the DocRED manual training and one on the ReDocRED manual training dataset. \citet{menotti_etal-2026-DOREMI} exploits an active learning pipeline to perform data denoising tailored for long-tail relations leveraging a set of \ac{DocRE} models. As for UGDRE, this approach generated two distant datasets, one exploiting the DocRED manual training and the other using ReDocRED.  To our knowledge, all denoising techniques for \ac{DocRE} exploit the predictions of the model and focus on semantic quality rather than syntactic accuracy.

Logical rules have been explored in \ac{DocRE} to enhance relation prediction. \citet{ru_etal-2021} introduced a probabilistic framework that learns logic rules as latent variables and optimizes them via \ac{EM}. \citet{zhang_etal-2024} proposed a Secondary Reasoning Framework that leverages first-stage predictions as contextual signals to infer additional relations among neighboring entities. \citet{zhang_etal-2025c} developed GREP, which models interdependencies between entity pairs to capture fine-grained reasoning patterns.
These approaches incorporate logical or reasoning mechanisms within the prediction process, focusing on improving semantic inference. In contrast, our work addresses a complementary dimension: the structural well-formedness of triples under ontology-imposed constraints, independently of the model architecture or inference strategy.

Constraint-based approaches have also been explored in sentence-level \ac{RE} to improve performance under distant supervision. \citet{liu_etal-2014} leveraged entity type information to introduce fine-grained type constraints for relation prediction. \citet{ji_etal-2017} incorporated Freebase entity descriptions as background knowledge to guide attention mechanisms and improve extraction accuracy. \citet{liang_etal-2023} proposed a \emph{constraint graph} constructed from entity type constraints, which provides an additional supervision signal within a constraint-aware attention module.
While these methods integrate external knowledge to guide sentence-level inference, they have not been extended to \ac{DocRE}, nor do they explicitly enforce ontology-level structural consistency. %over extracted triples.

%\lm{Prior works on consistency-based data quality? Io toglierei tutta questa parte su \citet{cima_etal-2025} è un lavoro totalmente diverso dal nostro, anche se usano le ontologie. Loro fanno metadata analysis e cercano di capire se l'ontologia considerata è troppo strict o va bene per il dato database. Fatemi sapere se siete d'accordo e se per voi i related vanno bene così o aggiungiamo. \citet{cima_etal-2025} proposes a framework to assess the dimension of consistency in data quality based on \ac{OBDM} and tailored for metadata analysis. Consistency is defined as the ``\emph{property of data that is free of contradictions and incoherence}''. Ontologies are chosen for their well-defined formal semantics and because they are usually curated by human experts, making them high-quality artifacts.}

\section{Methodology}
\label{sec:rules}
% \lm{Description of all the syntactic aspects. For each aspect, we provide a definition and how we build the rules.}
%Given a document $d$ and a set of entities $E=\{e_i\}_{i=1}^n$ that can occur multiple times in the document (entity $e_i \in E$ has entity mentions $\{m_j^i\}_{j=1}^{N_{e_i}}$), the task of \ac{DocRE} is to predict a subset of relations from $\mathcal{R} \cup \{NA\}$ between the entity pairs $(e_s, e_o)_{s,o=1..,n;s \neq o}$. $\mathcal{R}$ is a pre-defined set of relations and $e_s,e_o$ are defined as subject and object entities, respectively. 
Given a document $d$ and a set of entities $E = \{e_i\}_{i=1}^n$, each entity $e_i \in E$ is associated with a set of mentions $M_i = \{m_{i,j}\}_{j=1}^{k_i}$, where $k_i$ denotes the number of occurrences of $e_i$ in $d$. 
The task of \ac{DocRE} is to predict a subset of relations from $\mathcal{R} \cup \{\texttt{na}\}$ between entity pairs $(e_s, e_o)$ for $s, o \in \{1, \dots, n\}$ and $s \neq o$. Here, $\mathcal{R}$ is a pre-defined set of semantic relations, $\texttt{na}$ represents the absence of a relation, and $e_s, e_o$ denote the subject and object entities.~\footnote{\add{We adopt the DocRED convention of \emph{subject} and \emph{object} to denote the first and second arguments of an RDF triple ⟨s,p,o⟩, not syntactic roles.}}
Rather than treating these predictions as independent classification decisions, we interpret \ac{DocRE} as the construction of a \ac{KG} focusing on local well-formedness conditions induced by \ac{OWL} axioms. This perspective allows us to detect structural inconsistencies that are invisible to standard training objectives. %In this work, we define structural consistency as the satisfaction of ontology-imposed constraints over relational triples, focusing on local well-formedness conditions induced by OWL axioms.

\subsection{Invalid Triples}

Each entity mention $m_{i,j}$ of an entity $e_i$ is classified into a specific type from the set of named entity types $\mathcal{T} = \{\text{PER, ORG, LOC, TIME, NUM, MISC}\}$.
For instance, the mention ``\emph{France}'' is classified as $\text{LOC} \in \mathcal{T}$. Since all mentions in a set, say $M_i$, identify the same entity $e_i$, we can assume consistent typing across the set; thus, we use the first mention as representative, i.e., $t(e_i) = t(m_{i,1})$.
%. Indeed, DocRED enforces consistent typing across mentions

%Each entity mention $m_j$ of entity $e_i$ in DocRED is classified by a named entity type, which can be: ``Person (PER)'', ``Organization (ORG)'', ``Location (LOC)'', ``Time (TIME)'', ``Number (NUM)'', and ``other types (MISC)''.  For instance, entity mention ``\emph{France}'' can be classified as ``Location (LOC)''. 

%Since all entity mentions $\{m_j^i\}_{j=1}^{N_{e_i}}$ identify the same entity $e_i$, we assume all entity mentions have the same named entity type. Thus, the entity type for $e_i$ is defined as the named entity type of the first entity mention ($m_0^i$).  

In this context, the entity pair $(e_s, e_o)$ and their predicted relation $r \in \mathcal{R}$ form a \ac{RDF} triple $\langle s, p, o\rangle$, where $s = e_s$ is the subject, $p = r$ is the predicate, and $o = e_o$ is the object.  Beyond simple triple extraction, we can leverage \ac{OWL} to define formal semantics over the set of entities $E$ and relations $\mathcal{R}$. Each relation $r \in \mathcal{R}$ can be viewed as an \texttt{owl:ObjectProperty} with specific domain and range constraints. Specifically, for a triple $\langle e_s, r, e_o\rangle$ to be semantically valid, the entity types must satisfy:
$t(e_s) \in \text{dom}(r) \quad \text{and} \quad t(e_o) \in \text{ran}(r)$, where $\text{dom}(r), \text{ran}(r) \subseteq \mathcal{T}$ are the permitted subject and object types for relation $r$. 

Triples that violate these constraints
%, such as a ``\textit{place\_of\_birth}'' relation existing between two entities of type \text{ORG}, 
are considered \textit{invalid triples} within the defined ontology. By treating $\mathcal{R}$ as a set of restricted properties rather than arbitrary labels, we can filter out noise and improve the logical consistency of the extracted document graph.

%We can draw a parallelism with RDF and consider the entities as instances (named individuals) and the relations as properties. Similarly, the entity types can be mapped to classes. For each relation, we can define its domain and range exploiting entity types. 
%To identify the domain and range of each relation, we leverage the ReDocRED training dataset and the wikidata relation descriptions. 
Each relation in DocRED is mapped to a wikidata property.~\footnote{For instance, metadata for property P26 (\emph{spouse}) is available at: \url{https://www.wikidata.org/wiki/Property:P26}.} Thus, we leverage the Wikidata property descriptions to identify the domain (metadata field ``\texttt{subject type constraint}'') and the range of each property (``\texttt{value type constraint}''), if present. We also use the ReDocRED training dataset~\cite{tan_etal-2022b} to extract the most popular entity types of the subject and object entities of each relation and include them in the domain and range. To avoid overly restrictive criteria, the domain and range of all relations include entity type ''MISC''. For instance, relation P19 (\emph{place of death}) has domain ``PER$|$MISC'' and range ``LOC$|$MISC''.

%\textbf{Invalid triple.} Given an entity pair $(e_s,e_o)$, relation $r$ is defined \emph{invalid} for the pair if the subject entity type or the object entity type is not included in the domain or range of the property. A triple \mbox{$<e_s,r,e_o>$} where relation $r$ is invalid for the pair $(e_s,e_o)$ is called an \emph{invalid triple}. For instance, <Emily (PER), P19, 1987 (TIME)> is an invalid triple because the object entity type (TIME) is not allowed for relation P19 (\emph{place of death}). The complete list of domain and range of each relation is reported in Tables~\ref{tab:domain-range-rules1} and~\ref{tab:domain-range-rules2} of Appendix~\ref{app:rules}.

\begin{definition}[Invalid Triple]
Let $e_s, e_o \in E$ be a pair of entities with associated types $t(e_s), t(e_o) \in \mathcal{T}$ and $r \in \mathcal{R}$ be a relation; $r$ is defined as \textbf{invalid} for the pair $(e_s, e_o)$ if the subject entity type or the object entity type is not included in the domain or range of the property: $t(e_s) \notin \text{dom}(r) \quad \lor \quad t(e_o) \notin \text{ran}(r)$.
\end{definition}

A triple $\langle e_s, r, e_o \rangle$ where relation $r$ is invalid for the pair $(e_s, e_o)$ is called an \textbf{invalid triple}. For instance, $\langle \text{Emily (PER)}, \text{P19}, \text{1987 (TIME)} \rangle$ is an invalid triple because the object entity type (\text{TIME}) is not allowed for relation P19 (\emph{place of death}), as $\text{TIME} \notin \text{ran}(\text{P19})$. The list of domain and range of each relation is reported in Tables~\ref{tab:domain-range-rules1} and~\ref{tab:domain-range-rules2} of Appendix~\ref{app:rules}.

% For each relation, we identify the domain and range of the relation. The domain of a relation is defined as the subset entity types allowed for the subject entity of the relation. Similarly, the range of a relation is defined as the object entity types allowed for the object entity of the relation. For instance, relation P19 (\emph{place of death}) has domain [PER, MISC] and range [LOC, MISC].

\subsection{Missing Inverse Relations}
%Inverse properties in RDF allow defining bidirectional relationships between instances. Similarly, we identify inverse relations for a subset of $23$ DocRED relations. For instance, relation P1376 (\emph{capital of}) is the inverse relation of P36 (\emph{capital}). To identify inverse relations in DocRED, we leverage the metadata field ``\texttt{inverse property}'' of the wikidata property descriptions. The complete list of inverse relations is reported in Table~\ref{tab:inverses} of Appendix~\ref{app:rules}. 

%\textbf{Missing inverse relation.} Given a triple \mbox{$<e_s,r,e_o>$}, where $r$ has inverse relation $p$, there must exist a triple \mbox{$<e_o,p,e_s>$}. We do not assume $r \neq p$ to include \emph{symmetric} relations.

Inverse properties in \ac{OWL} allow for the formal definition of bidirectional relationships between entity instances. Within each document-level extracted graph, we adopt a local closed-world assumption with respect to inverse relations: if a relation $r$ holds between two entities in the document, its declared inverse should also be present in the extracted graph. We identify a subset of relations $\mathcal{R}_{inv} \subset \mathcal{R}$ that possess an inverse property. Formally, for a relation $r \in \mathcal{R}$, there may exist a relation $r' \in \mathcal{R}$ such that $r' \equiv \texttt{owl:inverseOf} \, r$. For instance, the relation $P1376$ (\textit{capital of}) is the inverse of $P36$ (\textit{capital}). To identify these pairs, we leverage the ``\texttt{inverse property}'' metadata field from the Wikidata properties description.

%\begin{definition}[Missing Inverse Relation]
%Let $r, r' \in \mathcal{R}$ be two relations such that $r'$ is the \texttt{owl:inverseOf} $r$. In a logically consistent document $d$, for every triple $\langle e_s, r, e_o \rangle$, there must exist an inverse triple $\langle e_o, r', e_s \rangle$. 
%Hence, a pair of entities $(e_s, e_o)$ is said to have a \textbf{missing inverse relation} if: $\langle e_s, r, e_o \rangle \in d \quad \land \quad \langle e_o, r', e_s \rangle \notin d$. %, where $d$ is the document from which the triples are extracted. 
%\end{definition}

\begin{definition}[Missing Inverse Relation]
Let $r, r' \in \mathcal{R}$ be two relations such that $r'$ is declared as the \texttt{owl:inverseOf} $r$ in the ontology. By the definition of \texttt{owl:inverseOf}, the presence of a triple $\langle e_s, r, e_o \rangle$ in a document $d$ entails the existence of the inverse triple $\langle e_o, r', e_s \rangle$. Hence, a pair of entities $(e_s, e_o)$ is said to have a \textbf{missing inverse relation} if:
\[
\langle e_s, r, e_o \rangle \in d \quad \land \quad \langle e_o, r', e_s \rangle \notin d
\]
\end{definition}

Note that $r$ and $r'$ are not required to be distinct; if $r = r'$, the relation is defined as an \texttt{owl:SymmetricProperty}, where the predicate must hold in both directions. 
The list of inverse relations is reported in Table~\ref{tab:inverses} of Appendix~\ref{app:rules}.

%If a relation $r$ is not symmetric, then it is called asymmetric. We define $<e_s,r,e_o>$ to be an \emph{asymmetric triple} if the relation $r$ is asymmetric. Leveraging the metadata field ``\texttt{inverse property}'' of the wikidata property descriptions, we discover that DocRED comprises only two symmetric relations: P26 (\emph{spouse}) and P3373 (\emph{sibling}). All other relations are asymmetric.

%\textbf{Asymmetric relation violation.} Given an asymmetric triple \mbox{$<e_s,r,e_o>$}, there must not exist a triple \mbox{$<e_o,r,e_s>$}.

\subsection{Asymmetric Relation Inconsistencies}
A relation $r \in \mathcal{R}$ is considered asymmetric if the existence of a relationship in one direction precludes its existence in the opposite direction. Formally, leveraging the ``\texttt{inverse property}'' metadata from Wikidata, we identify that the DocRED dataset contains only two symmetric relations: $P26$ (\textit{spouse}) and $P3373$ (\textit{sibling}), both of which satisfy $r \equiv \texttt{owl:SymmetricProperty}$. All other relations $r \in \mathcal{R} \setminus \{P26, P3373\}$ are treated as asymmetric.

\begin{definition}[Asymmetric Relation Violation]
Let $r \in \mathcal{R}$ be an \texttt{owl:AsymmetricProperty}. A triple $\langle e_s, r, e_o \rangle$ is defined as an \textbf{asymmetric triple}. A document $d$ contains an \textbf{asymmetric relation violation} if both the triple and its direct inverse (using the same relation $r$) are present:
$    \langle e_s, r, e_o \rangle \in d \quad \land \quad \langle e_o, r, e_s \rangle \in d$, where $s \neq o$. Such a state represents a logical contradiction within the \ac{OWL} definition of asymmetry, as the property cannot be its own inverse unless it is symmetric.
\end{definition}

\subsection{Relations Cardinality Violations}

Given an entity subject $e_s$, the maximum cardinality of a relation $r$ is defined as the largest possible number of distinct entity objects \(e_o\) such that triples of the form \(\langle e_s, r, e_o \rangle\) can be instantiated.
While domain knowledge initially suggests that a subset of $11$ relations should be strictly functional ($k_r = 1$), empirical analysis of the ReDocRED training set reveals that $7\%$ of these relation instances exhibit a maximum cardinality of 2. This is often due to ground truth comprising both full dates (e.g., ``\textit{2 July 1931}'') and separate year mentions (e.g., ``\textit{1931}'') as distinct entities. For example, $15\%$ of training instances of $P570$ (\textit{date of death}) present $k_r > 1$. Consequently, to avoid overly restrictive criteria, we relax the maximum cardinality of such relations to $k_r = 2$. The complete mapping of relations to their respective maximum cardinalities is provided in Table~\ref{tab:cardinality} of Appendix~\ref{app:rules}.
%While domain knowledge initially suggests that certain relations should be strictly functional ($k_r = 1$), empirical analysis of the ReDocRED training set reveals that $7\%$ of such instances exhibit a cardinality of 2. This is often due to the ground truth labeling both a full date (e.g., ``\textit{2 July 1931}'') and a separate year mention (e.g., ``\textit{1931}'') as distinct entities, as for $P570$ (\textit{date of death}) where $15\%$ of its instances present $k_r > 1$. Consequently, we identify a subset of $11$ relations where $k_r = 2$. The complete mapping of relations to their respective maximum cardinalities is provided in Table~\ref{tab:cardinality} of Appendix~\ref{app:rules}.

\begin{definition}[Cardinality Violation]
Given a relation $r \in \mathcal{R}$ with maximum cardinality $k_r$, let $O(e_s, r)$ be the set of object entities associated with a subject $e_s$ and relation $r$ in the document $d$:  $O(e_s, r) = \{ e_o \in E \mid \langle e_s, r, e_o \rangle \in d, s \neq o \}$. A \textbf{cardinality violation} occurs for the subject $e_s$ if the number of distinct extracted triples exceeds the defined threshold: $|O(e_s, r)| > k_r$.
\end{definition} 

In \ac{OWL} terms, if $k_r = 1$, the relation $r$ is treated as an \texttt{owl:FunctionalProperty}; for $k_r > 1$, it is treated as a \texttt{qualified cardinality restriction}.

\section{Structural Consistency in DocRE}
\label{sec:data-quality}
We analyze the structural consistency of the DocRED datasets using the rules defined above. Given DocRED is the reference test collection for the task, we primarily employ the DocRED distant dataset. We perform the same experiments in the distant datasets produced by the two leading denoising strategies, namely UGDRE~\cite{sun_etal-2023} and DOREMI~\cite{menotti_etal-2026-DOREMI}. The results on denoised distant datasets are reported in Appendix~\ref{app:denoising}. 

%We focus our analysis on the DocRED dataset~\cite{yao_etal-2019} and its different versions~\cite{tan_etal-2022b,sun_etal-2023,menotti_etal-2026-DOREMI}. 
\add{We analyzed several DocRE datasets from different domains (especially biomedical), but none of them was suitable for the study. BioRED~\cite{luo_etal-2022} and BC5CDR~\cite{li_etal-2016} are two well-known biomedical DocRE datasets, but they only include one relation type (``\texttt{associated\_with}") between different entity types. While GutBrainIE~\cite{martinelli_etal-2026} includes a sufficient variety of entity types and relations, its preprocessing to remove structural inconsistencies before release renders our analysis infeasible.}

To compare the amount of noise in distantly and manually annotated data, we also consider the DocRED and ReDocRED manual training datasets. In addition, we perform the same analysis on the ReDocRED evaluation datasets (dev and test), which are the revised version of the split DocRED development dataset. For each dataset, we report the number of errors for each structural constraint (Section~\ref{subsec:quality}). The code to perform the quality assessment on DocRED and ReDocRED datasets is available online. We also perform a statistically grounded qualitative analysis on invalid triples in the DocRED distant dataset to understand the main cause of such errors (Section~\ref{subsec:sampling}). 

\begin{table*}[t!]
\resizebox{\textwidth}{!}{%
    \centering
    \begin{tabular}{l|l|l|llll}
    \toprule
    Dataset & Split & \multicolumn{1}{c|}{Triples} & \multicolumn{1}{c}{Invalid Triples ($\delta$)} & \multicolumn{1}{c}{Missing Inverse ($\sigma$)} & \multicolumn{1}{c}{Asymmetric Inc. ($\gamma$)} & \multicolumn{1}{c}{Cardinality Viol. ($\theta$)} \\
    \midrule
    DocRED & Train & 38,180 & 602 (1.58\%) & 2,696 (64.36\%) & 154 (0.41\%) & 10 (0.28\%) \\  
    ReDocRED & Train & 85,932 & 810 (0.94\%) & 947 (7.89\%) & 284 (0.34\%) & 20 (0.40\%) \\
    ReDocRED & Dev & 17,284 & 122 (0.71\%) & 243 (8.92\%) & 56 (0.33\%) & 3 (0.32\%) \\
    ReDocRED & Test & 17,448 & 141 (0.81\%) & 242 (9.27\%) & 82 (0.48\%) & 6 (0.65\%) \\
    \midrule
    DocRED & Train DS & 1,505,638 & 46,211 (3.07\%) & 85,065 (40.26\%) & 12,408 (0.84\%) & 25 (0.01\%) \\
    \bottomrule
    \end{tabular}%
}
\caption{Structural consistency analysis of DocRED datasets. Splits ``Train'', ``Dev'', and ``Test'' consider manually annotated datasets, while split ``Train DS'' reports the quality of the DocRED distant dataset. For each rule, we report the absolute number of errors in the dataset and its percentage. }
\label{tbl:data-quality}
\end{table*}
\subsection{Datasets Consistency Assessment}
\label{subsec:quality}
%\lm{Table with syntactic errors in each dataset. FINDING 1: distant datasets contains more syntactic inconsistencies than the manual datasets (quantify the number of errors). FINDING 2: Denoising techniques are ineffective on syntactic accuracy.}
Table~\ref{tbl:data-quality} reports the structural consistency of the DocRED datasets. If distant supervision introduces systematic structural degradation, we expect higher violations in the DocRED distant split compared to manual datasets. 
For each structural constraint defined in Section~\ref{sec:rules}, we report both the absolute number of violations and their percentage, computed according to the specific constraint type. 

The percentage of invalid triples ($\delta$) is defined as the ratio between the number of structurally invalid triples and the total number of triples in the dataset. The percentage of inverse triples missing in a dataset ($\sigma$) is the ratio between the number of missing inverse triples and the triples involving relations with inverses. The percentage of asymmetric inconsistencies in a dataset ($\gamma$) is the ratio between such inconsistencies and the triples involving an asymmetric relation. Finally, the percentage of cardinality violations ($\theta$) is the ratio between the subject entity-relation pair $(e_s,r)$ violating $|O(e_s,r)| > k_r$ over all such pairs involving a relation with bounded maximum cardinality.

As expected, DocRED distant presents much more inconsistencies than the manual datasets. The only exception is on the missing inverse relations, where the worst is the DocRED manual training dataset. This is a known issue in the literature and has been corrected in ReDocRED~\cite{huang_etal-2022,tan_etal-2022b}, as demonstrated by the reduction in missing inverse between the DocRED and ReDocRED manual training dataset (from around $64\%$ to less than $8\%$). Compared to the ReDocRED manual training dataset, the DocRED distant dataset contains three times more invalid triples, nearly five times more missing inverse relations, and twice as many asymmetric inconsistencies. About cardinality violations, DocRED distant is the dataset containing less errors ($0.01\%$), while the other datasets range between $0.28\%$ and $0.65\%$ of violations. This phenomenon can be attributed to the distant supervision methodology employed in constructing the DocRED distant dataset, which aligns Wikipedia documents with Wikidata triples and thereby preserves their cardinalities.

Also ReDocRED datasets exhibit structural violations, with a small but non-negligible proportion of errors in all splits. Invalid triples constitute between $0.71\%$ and $0.94\%$ of the datasets, missing inverse relations remain below $10\%$, and asymmetric violations range from $0.33\%$ to $0.48\%$. Cardinality violations exceed those observed in the DocRED manual training and DocRED distant, reaching $0.65\%$ in the ReDocRED test set. Given the presence of structural violations in the evaluation datasets, models trained to generate well-formed triples may exhibit degraded performance.

\subsection{Invalid Triples Analysis}
\label{subsec:sampling}

To identify the underlying causes of invalid triples in the distant dataset, we conduct a statistically grounded error analysis. We classify each invalid triple into one of two mutually exclusive categories: (i) relation annotation errors, where, given a correctly typed entity pair, the wrong relation type is assigned, and (ii) entity annotation errors, where one or both entities are incorrectly typed, thereby inducing an invalid relation. 
For example, the triple $\mathtt{\langle \textit{France} (LOC), P16 (\textit{capital}), \textit{Paris} (PER)\rangle}$ is considered an entity annotation error because ``\emph{Paris}'' is mistakenly classified as a person.% rather than a location.

Since annotating all invalid triples within DocRED distant would be infeasible in terms of time and cost, we instead rely on binomial proportion estimation techniques~\cite{brown_etal-2001}. This approach enables us to obtain statistically reliable estimates while limiting the amount of manual annotation required. The evaluation protocol uses unbiased sampling and estimation procedures, providing strong guarantees on the robustness of the reported results. 

Specifically, we draw a sample using \ac{SRS} of size $n_{\mathcal{S}} = 400$ from the DocRED distant dataset population of invalid triples, ensuring representativeness of the sample~\cite{cochran-1977}. Each sampled triple is manually annotated according to the error classification described above. We formalize the annotation process as a binomial indicator function $\mathds{1}(\langle e_s, r, e_o \rangle)$, which takes value $1$ if the triple contains an entity annotation error (e.g., incorrect entity type) and $0$ if it contains a relation annotation error. %For example, the invalid triple $\langle \textit{France} (LOC), P16 (\textit{capital}), \textit{Paris} (PER)\rangle$ is assigned value $1$, since the entity ``Paris'' is incorrectly labeled as a person.

After annotation, we compute the sample proportion $\hat{\mu} = \frac{\sum_{\langle e_s, r, e_o \rangle\in \mathcal{S}} \mathds{1}(\langle e_s, r, e_o \rangle)}{n_{\mathcal{S}}}$, which is an unbiased estimator of the true population proportion~\cite{cochran-1977}. To quantify sampling uncertainty, we accompany this estimate with a $1-\alpha$ credible interval. We employ Highest Posterior Density (HPD) intervals~\cite{box_tiao-2011}, which are well suited for uncertainty quantification in \ac{KG} evaluation settings~\cite{marchesin_silvello-2025}. Unlike (frequentist) confidence intervals, HPD intervals operate directly in the probabilistic space and avoid common interpretational pitfalls~\cite{morey_etal-2016}, making them appropriate for one-shot evaluation scenarios~\cite{marchesin_silvello-2025}. We set $\alpha=0.05$, yielding a $95\%$ credible interval.

%Building on this setup, we perform a manual annotation step to estimate how much each of the two error sources contributes to the overall pool of invalid triples. We randomly select $400$ invalid triples from DocRED distant and manually annotate each one according to the error taxonomy introduced above. Based on these annotations, we estimate the share of invalid triples attributable to entity errors versus relation errors, and we report a $95\%$ credible interval to capture sampling uncertainty. The evaluation protocol uses unbiased sampling and estimation procedures, providing strong guarantees on the robustness of the reported results. 

%A comprehensive account of the statistical error analysis can be found in Appendix~\ref{app:sampling}.

The analysis shows that $69\%$ ($\pm 4.5\%$) of invalid triples stem from errors in entity annotation, while the remaining $31\%$ are attributable to incorrect relation annotations. This pronounced disparity demonstrates that entity-related mistakes are the dominant source of invalid triples. As a result, the findings imply that enhancing named entity recognition and employing targeted denoising methods for entity annotations could markedly improve the structural reliability of distant supervision datasets. %More generally, the results underscore substantial untapped potential for advances in entity recognition -- an area that, despite its central importance for downstream information extraction, has attracted relatively limited attention in recent years.
More broadly, these findings emphasize that entity extraction is a bottleneck for relation extraction. In end-to-end pipelines, inaccuracies in entity recognition propagate to relation prediction, thereby imposing an upper bound on the overall performance of \ac{RE} systems. Consequently, improvements in relation extraction are inherently capped by the quality of entity annotations.
In this work, we do not modify or refine entity extraction; our intervention is limited to enforcing structural consistency over the extracted triples. As a result, the performance gains we observe should be interpreted as conservative estimates: further improvements could be achieved by jointly addressing entity-level errors. Moreover, since entity misannotations are also present in the evaluation data, correcting structurally inconsistent relations may correspond to semantically correct predictions in real-world settings, yet still be penalized under the current benchmark due to pre-existing entity annotation errors.

\section{Training Effect}
\label{sec:training}
We investigate if training \ac{DocRE} models with structurally consistent data improves their performance. 

\subsection{Cleaning the datasets}
\label{sub:cleaning}
% \lm{Description of how we correct each type of syntactic error. Report the statistics of the corrected distant datasets. Share the notebook to clean any dataset.}
%We clean the DocRED distant dataset and generate a distant dataset free of structural errors. Given the minimal presence of cardinality inconsistencies in all the datasets, we decide not to clean such errors and only consider invalid triples, missing inverse relations, and asymmetric violations. 
%Correcting invalid triples implies a manual review of the entities and relations. Due to the high number of invalid triples in the DocRED distant dataset ($46,211$), manually correcting all invalid triples is infeasible in terms of annotation cost. Thus, instead of correcting the triples, we remove all entities involved in at least an invalid triple from the training dataset. This operation resulted in the removal of $208,665$ triples and $3$ documents.  
We construct a structurally cleaned version of the DocRED distant dataset by eliminating triples that violate the ontological constraints defined in Section~\ref{sec:rules}. Since cardinality violations are extremely rare across all splits, especially in DocRED distant ($0.01\%$), we focus our cleaning procedure on invalid triples, missing inverse relations, and asymmetric violations. 

Correcting invalid triples would require manual inspection and revision of both entity types and relation labels. Given the large number of invalid triples in the DocRED distant dataset ($46{,}211$), fully correcting them would entail extensive manual effort and substantial annotation cost. Instead, we adopt a conservative and scalable strategy: for each invalid triple, we remove the entity involved in the violation from the training set, thereby eliminating all triples associated with that entity within the document. 
% This guarantees the removal of structurally inconsistent triples without introducing additional annotation noise.
Under this strategy, removing an entity effectively removes the offending triple(s) that depend on its incorrect typing or relation assignment. This approach is significantly cheaper than manual correction, as it avoids entity re-labeling while ensuring that no structurally invalid triples remain in the cleaned dataset. Overall, this procedure results in the removal of $208{,}665$ triples and $3$ documents from DocRED distant.

Missing inverse relations are added automatically, resulting in the addition of $72,609$ triples.
Similarly to invalid triples, solving asymmetric inconsistencies requires the manual evaluation of $12,408$ triples. Thus, we remove all entities involved in at least one asymmetric violation from the dataset, resulting in the removal of $31,178$ triples. 

As a result, the \emph{corrected} distant dataset contains $1,338,404$ training instance ($-167,234$ than DocRED distant) and $101,870$ documents ($-3$ than DocRED distant). The code to correct structural inconsistencies in \ac{DocRE} datasets is available in the online repository. The code is compatible with any dataset in DocRED format. The repository contains the rules specifically developed for DocRED, but the code can also run on user-defined constraints. 

\subsection{Experimental Setup}
\label{sub:setup}
%\lm{Description of how we check the effect during trianing. Explanation of the model of choice (ALTOP and DREEAM) and the list of the datasets. NBB: When using DREEAM, the teacher model is always the one trained on ReDocRED manual.}
To explore the effect of structural violations on training, we leverage two \ac{DocRE} models with leading performance and open-source implementations: ATLOP~\cite{zhou_etal-2021} and DREEAM~\cite{ma_etal-2023}. We chose two transformer-based models because graph-based models cannot scale to the size of distant datasets. ATLOP exploits a BERT encoder to represent entity pairs enriched with localized context pooling. During prediction, adaptive thresholding is introduced to better handle multi-label classification~\cite{zhou_etal-2021}. DREEAM is a teacher-student model, whose architecture is based on ATLOP enriched with an evidence-aware distillation mechanism~\cite{ma_etal-2023}. We train both models on the DocRED distant dataset and its \emph{corrected} version, using the ReDocRED development dataset for validation. DREEAM follows a teacher–student paradigm, where the student model follows a first step of ``\emph{self-training}'' and then is fine-tuned on manual data. During self-training, the student model is trained on distant data and weakly supervised signals from the teacher model. Since our focus is on evaluating the quality of DocRED distant, we consider the DREEAM student model right after self-training, exploiting the teacher model trained on the ReDocRED training dataset.  We then evaluate each model on the ReDocRED test dataset and exploit their predictions on these documents to assess the number of ill-formed triples. The code to analyze the structural consistency of the predictions is available online. The code is compatible with any \ac{DocRE} model predictions formatted as DocRED official results and can be used with any user-defined rules.

\begin{table}[t!]
\resizebox{\columnwidth}{!}{%
    \centering
    \begin{tabular}{c|c|l|llll}
    \toprule
    & PLM & Training Data & \multicolumn{1}{c}{Invalid ($\delta$)} & \multicolumn{1}{c}{Inverse ($\sigma$)} & \multicolumn{1}{c}{Asymmetric ($\gamma$)} \\
    \midrule
    \multirow{4}{*}{\rotatebox[origin=c]{90}{ATLOP}} & \multirow{2}{*}{BERT} & DocRED DS & 110 (1.61\%) &  396 (43.04\%) & 38 (0.56\%) \\
    & & \cellcolor{ours} Corrected & \cellcolor{ours} \textbf{46 (0.66\%)} & \cellcolor{ours} \textbf{65 (5.61\%)} & \cellcolor{ours} \textbf{4 (0.06\%)} \\
    \cmidrule{2-6}
    & \multirow{2}{*}{RoBERTa} & DocRED DS & 110 (1.56\%) & 407 (45.22\%) & 42 (0.60\%) \\
    & & \cellcolor{ours} Corrected & \cellcolor{ours} \textbf{42 (0.59\%)} & \cellcolor{ours} \textbf{66 (5.68\%)} & \cellcolor{ours} \textbf{0 (0.00\%)} \\
    \specialrule{1pt}{0pt}{2pt}
    \multirow{4}{*}{\rotatebox[origin=c]{90}{DREEAM}} & \multirow{2}{*}{BERT} & DocRED DS & 59 (1.05\%) & 266 (41.18\%) & 30 (0.53\%) \\
    & & \cellcolor{ours} Corrected & \cellcolor{ours} \textbf{21 (0.31\%)} & \cellcolor{ours} \textbf{115 (11.97\%)} & \cellcolor{ours} \textbf{4 (0.06\%)} \\
    \cmidrule{2-6}
    & \multirow{2}{*}{RoBERTa} & DocRED DS & 52 (0.85\%) & 333 (44.76\%) & 48 (0.79\%) \\
    & & \cellcolor{ours} Corrected & \cellcolor{ours} \textbf{23 (0.34\%)} & \cellcolor{ours} \textbf{92 (10.15\%}) & \cellcolor{ours} \textbf{6 (0.09\%)} \\
    \bottomrule
    \end{tabular}%
}
\caption{Structural consistency of DocRE models predictions trained on DocRED Distant (``\emph{DocRED DS}'') and its structurally-consistent version (``\emph{Corrected}'') and evaluated on \textbf{ReDocRED test dataset}. For each rule, we report the absolute number of errors in the predictions and its percentage. Column ``\emph{Invalid}'' refers to invalid triples, ``\emph{Inverse}'' to missing inverse triples, and ``\emph{Asymmetric}'' to asymmetric inconsistencies.}
\label{tbl:train-preds}
\end{table}

\subsection{Structural Consistency in predictions}
\label{sub:acc-preds}
%\lm{Table with the number of syntactic errors in the predictions of the model trained with the different datasets. FINDING: training with syntactically accurate data leads to syntactically accurate predictions.}
Table~\ref{tbl:train-preds} reports the structural analysis of predictions by ATLOP and DREEAM models, trained on DocRED distant and its \emph{structurally-consistent} version, all evaluated on ReDocRED test set. 

% Train DS Invalid 3.07% Inverse 40.26% Asymmetric 0.84%
Training DocRE models on structurally inconsistent data results in noisy predictions, with ATLOP appearing slightly more sensitive than DREEAM. The greater robustness of DREEAM may stem from its self-training procedure, which combines distant data with weak supervision signals provided by a teacher model trained on manual data. The proportion of invalid triples in the predictions ranges from $0.85\%$ to $1.61\%$, about one third of the rate observed in the training data. Missing inverse triples in the training set propagate almost linearly to the predictions: while $40.6\%$ of inverse triples are missing in the dataset, the corresponding figures in the predictions range from $41.18\%$ to $45.22\%$. A similar pattern holds for asymmetric violations, where the $0.84\%$ rate observed in the training data closely matches the $0.53\%-0.79\%$ range found in the predictions.

In contrast, training \ac{DocRE} models without logical inconsistencies consistently improves their ability to generate well-formed triples. Also in this case, the improvement is more pronounced for ATLOP than for DREEAM. %, likely because DREEAM's teacher model already mitigates part of the structural noise. 
On average, the proportion of invalid triples is reduced by $-62.92\%$. The most substantial improvement is observed in inverse relation prediction: for ATLOP-RoBERTa, the percentage of missing inverse relations drops by $-87.44\%$, and for ATLOP-BERT by $-86.97\%$. DREEAM also benefits, with a reduction of $-74.13\%$. In both architectures, asymmetric violations are nearly eliminated; Training ATLOP-RoBERTa on structurally consistent data reduces asymmetric violations from $0.60\%$ to $0.00\%$. 

\begin{table*}[t!]
\resizebox{\textwidth}{!}{%
    \centering
    \begin{tabular}{c|c|l|lllll}
    \toprule
    & PLM & Training Data & Precision & IgnPrec & Recall & F1 & IgnF1 \\
    \midrule
    \multirow{6}{*}{\rotatebox[origin=c]{90}{ATLOP}} & \multirow{3}{*}{BERT} & DocRED DS & 0.7503 & 0.5877 & 0.2945 & 0.4230 & 0.3924 \\
    \cmidrule{3-8}
    & & \cellcolor{ours} Corrected & \cellcolor{ours} \textbf{0.7665} \inp{2.16} & \cellcolor{ours} \textbf{0.6165} \inp{4.90} & \cellcolor{ours} \textbf{0.3070} \inp{4.24} & \cellcolor{ours} \textbf{0.4384} \inp{3.64} & \cellcolor{ours} \textbf{0.4099} \inp{4.46} \\
    & & Random & 0.7528 \inp{0.33} & 0.5973 \inp{1.63} & 0.2964 \inp{0.65} & 0.4253 \inp{0.43} & 0.3962 \inp{0.97} \\
    \cmidrule{2-8}
    & \multirow{3}{*}{RoBERTa} & DocRED DS & 0.7571 & 0.6006 & 0.3060 & 0.4358 & 0.4054 \\
    \cmidrule{3-8}
    & & \cellcolor{ours} Corrected & \cellcolor{ours} \textbf{0.7657} \inp{1.14} & \cellcolor{ours} \textbf{0.6172} \inp{2.76} & \cellcolor{ours} \textbf{0.3120} \inp{1.96} & \cellcolor{ours} \textbf{0.4434} \inp{1.74} & \cellcolor{ours} \textbf{0.4245} \inp{4.71} \\
    & & Random & 0.7578 \inp{0.09} & 0.6082 \inp{1.27} & 0.3049 \dep{0.36} & 0.4348 \dep{0.23} & 0.4061 \inp{0.17}\\
    \specialrule{1pt}{0pt}{2pt}
    \multirow{6}{*}{\rotatebox[origin=c]{90}{DREEAM}} & \multirow{3}{*}{BERT} & DocRED DS & \textbf{0.8087} & 0.6791 & 0.2609 & 0.3945 & 0.3770 \\
    \cmidrule{3-8}
    & & \cellcolor{ours} Corrected & \cellcolor{ours} 0.8042 \dep{0.56} & \cellcolor{ours} \textbf{0.6941} \inp{2.21} & \cellcolor{ours} \textbf{0.3113} \inp{19.32} & \cellcolor{ours} \textbf{0.4488} \inp{13.76} & \cellcolor{ours} \textbf{0.4298} \inp{14.01} \\
    & & Random & 0.7974 \dep{1.40} & 0.6746 \dep{0.66} & 0.2683 \inp{2.84} & 0.4013 \inp{1.72} & 0.3837 \inp{1.78} \\
    \cmidrule{2-8}
    & \multirow{3}{*}{RoBERTa} & DocRED DS & \textbf{0.8170} & 0.6964 & 0.2861 & 0.4238 & 0.4056 \\
    \cmidrule{3-8}
    & & \cellcolor{ours} Corrected & \cellcolor{ours} 0.8143 \dep{0.33} & \cellcolor{ours} \textbf{0.7083} \inp{1.71} & \cellcolor{ours} \textbf{0.3113} \inp{8.81} & \cellcolor{ours} \textbf{0.4504} \inp{6.28} & \cellcolor{ours} \textbf{0.4353} \inp{7.32} \\
    & & Random & 0.8087 \dep{1.02} & 0.6974 \inp{0.14} & 0.2897 \inp{1.26} & 0.4263 \inp{0.59} & 0.4090 \inp{0.84} \\
    \bottomrule
    \end{tabular}%
}
\caption{Models trained on DocRED distant (``DocRED DS'') and its structurally-consistent version (``Corrected'') and evaluated on the \textbf{ReDocRED test dataset}. Random removal (``Random'') was performed ten times with different random seed and the mean performance is reported. The standard deviation is always below 1\%. The percentage performance improvement compared to DocRED distant is reported between parentheses.}
\label{tbl:train-eval}
\end{table*}
\subsection{Effect on model performance}
\label{sub:performance}
% \lm{Table with performance on full dataset for the model trained with the different datasets. FINDING: training with syntactically accurate data leads to improved performance.}
Having established that structural correction reduces ill-formed predictions, we next examine whether these structural improvements translate into better generalization performance. Table~\ref{tbl:train-eval} reports the results of \ac{DocRE} models trained on DocRED distant (row ``DocRED DS'') and its \emph{structurally-consistent} version (row ``Corrected''). \add{We investigate the impact of each noise source in Appendix~\ref{app:ablation}.} We recall that the corrected dataset contains $167,234$ fewer triples than DocRED distant. To ensure that any performance differences are not merely due to this reduction in training data, we additionally trained the models on a randomly subsampled variant obtained by removing the same number of triples ($167,234$) from DocRED distant. This random subsampling was repeated with ten different seeds, and the mean performance is reported in the ``Random'' row. To avoid clutter, Table~\ref{tbl:train-eval} does not report the standard deviation, as it is always below $1\%$. All models were validated on the ReDocRED development set and evaluated on the ReDocRED test set. \add{Together with precision, recall, and F1 scores, we also ignored metrics, that measures the precision (\textsc{ignPrec}) or F1 (\textsc{ignF1}) excluding the entity pairs seen during training.}

Training ATLOP on the corrected distant dataset improves performance across all metrics and both transformer backbones. Relative to DocRED distant, structural correction yields average gains of $+1.65\%$ in precision, $+3.10\%$ in recall, and $+2.69\%$ in F1. The improvements are even more pronounced under ignored metrics, where entity pairs seen during training are excluded. On average, \textsc{ignPrec} increases by $+3.83$, and \textsc{ignF1} by $+4.59\%$. 

DREEAM exhibits a similar overall trend, with improvements in all metrics except precision. Although precision decreases slightly (by $-0.45\%$ on average), this drop is compensated by major gains in the other metrics. Recall improves by $+14.07\%$ on average, with DREEAM-BERT achieving a boost of $+19.32\%$, while F1 rises by $+10.02\%$ on average. Under the ignored setting, \textsc{ignPrec} increases by $+1.96\%$ and \textsc{ignF1} by $+10.67\%$. 

%In contrast, r
Random removal of triples leads to only marginal improvements in most metrics and even degrades performance in several cases: recall and F1 for ATLOP-RoBERTa, precision for both DREEAM variants, and \textsc{ignPrec} for DREEAM-BERT. These findings indicate that the observed performance gains stem from the correction of structural errors rather than from the mere reduction in the number of training triples or documents.

\section{Conclusions}
\label{sec:conclusion}
% We introduced \ac{DOREMI}, a dataset enhancement framework that targets long-tail relation prediction while minimizing manual annotation. Leveraging iterative training, \ac{DOREMI} identifies Hard-To-Classify examples by measuring disagreement across multiple models. Being optimized for long-tail predictions, \ac{DOREMI} can complement existing denoising approaches, such as UGDRE, which are better suited for frequent relations. The resulting denoised distantly supervised dataset can be used to train any off-the-shelf \ac{DocRE} model, yielding improved performance on long-tail relation prediction. Experiments on ReDocRED show that \ac{DOREMI} significantly improves performance, especially in long-tail settings. Remarkably, annotating just $0.003\%$ of the DocRED \ac{DS} dataset yields F1 improvements of up to a $+56.7\%$ overall and $+53.2\%$ in long-tail predictions over UGDRE -- the current state-of-the-art in label denoising. DOREMI also significantly boosts long-tail \textsc{IgnF1} by $+59.5\%$, indicating better generalization to novel entity pairs unseen during training. These results highlight the effectiveness of disagreement-driven annotation -- enabling better generalization with negligible human effort.
We introduced a framework for assessing the structural consistency of \ac{DocRE} datasets and analyzed the impact of structural noise on model predictions and downstream performance. We formalized the \ac{DocRE} task as the construction of a \ac{KG} under the \ac{RDF} model and defined a set of logical constraints in \ac{OWL} to detect inconsistencies. In particular, we identified invalid triples through domain and range constraints on object properties; missing inverse relations via \texttt{owl:inverseOf}; asymmetric violations using \texttt{owl:AsymmetricProperty}; and relation cardinality violations through \texttt{qualified cardinality restrictions}. 

Our analysis shows that, compared to the ReDocRED manual training set, the DocRED distant dataset contains three times more invalid triples, nearly five times more missing inverse relations, and twice as many asymmetric inconsistencies. 
%About cardinality violations, DocRED distant is the dataset containing fewer errors ($0.01\%$), most likely due to the distant supervision methodology employed in constructing the dataset. 
To better understand the cause of invalid triples, we
conducted a qualitative analysis on a representative sample. Manual annotation revealed that almost $70\%$ of invalid triples stem from named entity misannotations. This finding suggests that effective denoising strategies should not only target relation labels but also explicitly address entity-level errors. 

Finally, we examined whether removing structural inconsistencies from the training data reduces ill-formed predictions and improves model performance. Training \ac{DocRE} models on structurally consistent data systematically lowers the number of structural violations in the predictions and leads to improved performance across nearly all metrics, with an average gain of $+6.36\%$ in F1 and $+7.62\%$ in \textsc{ignF1}. These results highlight the importance of structural data quality for both the reliability and the generalization ability of \ac{DocRE} systems.

\section{Limitations}
%\lm{TODO}

To evaluate the performance gains resulting from structural noise removal, we rely on the ReDocRED test set, which itself contains (a limited number of) logical inconsistencies. An ideal evaluation benchmark for isolating the effects of ill-formed training triples would be entirely free of structural noise.
However, re-annotating the ReDocRED test set could raise concerns about annotation bias, as such modifications might be perceived as being tailored to favor our approach.

Our noise mitigation strategy prioritizes entity removal rather than direct error correction. Correcting all ill-formed triples would require extensive manual revision of $58,619$ instances (including invalid triples and asymmetric inconsistencies), which would be prohibitively expensive in terms of time and resources. Instead, we chose to discard the affected entities, ensuring structural consistency while avoiding large-scale manual intervention. Future work could explore systematic re-annotation efforts to preserve a greater portion of the data while further improving dataset quality.

\add{Since we interpret the performance gains from structural correction as the removal of an irreducible noise term in the supervision signal, our error analysis is primarily empirical and does not yet quantify the statistical significance of the observed improvements across different sources of structural violation. In future work, we plan to complement our empirical findings with a more rigorous treatment, for example by deriving generalization bounds under structurally corrected labels or by analyzing the impact of constraint enforcement on the effective capacity of the hypothesis class.}

%\add{ACTION POINT: We will revise the limitations sections to address potential issues with the quality of the ontology definitions regarding incompleteness, noise, and over-generalizations.}

% Portato in sezione 3
%\add{We focus our analysis on the DocRED dataset~\cite{yao_etal-2019} and its different versions~\cite{tan_etal-2022b,sun_etal-2023,menotti_etal-2026-DOREMI}. We analyzed several DocRE datasets from different domains (especially biomedical), but none of them was suitable for the study. BioRED~\cite{luo_etal-2022} and BC5CDR~\cite{li_etal-2016} are two well-known biomedical DocRE datasets, but they only include one relation type (``\texttt{associated\_with}") between different entity types. While GutBrainIE~\cite{martinelli_etal-2026} includes a sufficient variety of entity types and relations, its preprocessing to remove structural inconsistencies before release renders our analysis infeasible.}

Finally, our analysis of training effects is restricted to two transformer-based DocRE architectures, ATLOP and DREEAM. Although these models represent strong and widely adopted baselines, future research could extend this investigation to graph-based and generative approaches. To support such efforts, we release our code and the full set of structural rules as open-source resources, enabling the community to reproduce and expand our structural consistency analysis across a broader range of DocRE systems.

\add{On the sample size for the qualitative error analysis, $n_{\mathcal{S}}=400$ yields a $95\%$ \ac{HPD} interval of $\pm 4.5\%$ around the $0.69$ entity-error proportion (Section~\ref{subsec:sampling} ). Halving the interval width requires $n_{\mathcal{S}} \approx 1600$, a four-fold annotation cost without altering the qualitative dominance of entity errors.}
%ACTION POINT: We will make this trade-off explicit in the limitations.}
%\ac{DOREMI} relies on aggregating labels predicted by iteratively trained models to produce a \ac{DDS}. In the multi-label context of DocRE, devising an effective label aggregation strategy is non-trivial and crucial for the quality of the resulting dataset. Future works could explore more sophisticated aggregation methods to enhance this process.

%The human annotation process in this study focused on determining the presence of relations between entity pairs, without verifying the correctness of the entity annotations themselves. Given that the entity annotations in the DocRED \ac{DS} are generated using a BERT-based \ac{NER} module, incorporating denoising mechanisms for entity annotations, in addition to relation labels, may further reduce dataset noise.

%The quality of the denoised datasets was evaluated indirectly by training state-of-the-art \ac{DocRE} models on them and measuring their performance on DocRED/ReDocRED development/test dataset. While this approach provides a useful signal, model performance serves only as a proxy for the true dataset quality and may not fully reflect the accuracy or consistency of the labels. High performance could arise from model robustness rather than genuinely clean data. Given the impracticality of large-scale manual validation in \ac{DS} settings, alternative methods -- such as accuracy estimation techniques commonly used in knowledge graph evaluation~\cite{ojha_talukdar-2017,gao_etal-2019} -- may offer more principled approximations of dataset quality. 

\section{Ethical Considerations}
Based on our methodology, we do not anticipate any significant ethical concerns. We list some potential ethical aspects and risks of our analysis:
\begin{itemize}

\item \textbf{Licenses and Data Privacy}: The dataset and models used in this work are open-sourced to minimize the risk of privacy leakage and ensure transparency and accessibility.
The \ac{DocRE} models used in the empirical evaluation are trained and evaluated exclusively on MIT licensed datasets, namely DocRED, ReDocRED, UGDRE, and DOREMI.
%The documents and models used in this study are sourced from open-source domains, ensuring transparency and accessibility. Using open-source datasets minimizes the risk of privacy leakage. 
%In addition, all \ac{DocRE} models exploited in the experimental evaluation are trained and evaluated exclusively on open source (e.g., MIT license) document-level relation extraction datasets, namely DocRED, ReDocRED, UGDRE, and DOREMI.   

\item \textbf{Hidden Biases}: The leading \ac{DocRE} models exploited for empirical analysis (ATLOP and DREEAM) are transformer-based models, relying on pre-trained language models such as BERT and RoBERTa. We recognize that pre-trained language models may include hidden biases from their training data, potentially embedding subtle human prejudices. While they perform well in detecting relations, special caution is needed for sensitive relation types or entities, where such biases may become more evident.

%ATLOP and DREEAM, the leading \ac{DocRE} models exploited for empirical analysis, rely on pre-trained language models such as BERT and RoBERTa. We acknowledge potential hidden biases in these pre-trained language models, which may stem from their training data and might embed subtle human prejudices. Although they excel at detecting relations between diverse entities, extra care is needed for sensitive relation types or entities, where biases could surface more prominently.

\item \textbf{The use of AI Assistants}: ChatGPT, Perplexity, Writefull, and Grammarly were used purely with the language of the paper to help improve clarity. All scientific content was developed by the authors, and the manuscript has been carefully reviewed and approved in its entirety by all authors.
\end{itemize}

%\textit{Based on the methodology we have currently employed, we do not foresee any significant ethical concerns. All the documents and models utilized in our study were obtained from open-source domains, ensuring a transparent and accessible source of information. Additionally, TTM-RE is trained on purely open-source document relation extraction data, eliminating the risk of privacy leakage. However, it is crucial to acknowledge a minor factor, namely the presence of potential hidden biases within the pre-trained language models used in our analysis. These biases may stem from the data on which the models were trained, which could have inadvertently introduced implicit human biases. While our usage of these pre-trained language models enables us to identify relationships between arbitrary entities, it is conceivable that biases may emerge if one were to explore sensitive relation classes and entities. ChatGPT and Grammarly were used for parts of the writing. In total, training took more than 75 hrs on NVIDIA RTX A6000 for pertaining in total. The main roadblock was the distantly supervised finetuning portion for all of the models, due to the size of the dataset. Derivatives of data accessed for research purposes should not be used outside of research contexts. Code will be released at https://github.com/chufangao/TTM-RE.}

\section*{Acknowledgments}

This project has received funding from the HEREDITARY Project, as part of the European Union's Horizon Europe research and innovation programme under grant agreement No GA 101137074. 

\bibliography{custom}
\newpage
\appendix
\section{Denoised Datasets}
\label{app:denoising}
We perform the same structural consistency analysis presented for DocRED distant on four denoised versions of the distant dataset, generated by two top-performing denoising techniques. UGDRE is a label denoising strategy that exploits Monte Carlo dropout to estimate pseudo-label uncertainty and filter out low-confidence triples produced by a \ac{DocRE} model trained on the manual training data~\cite{sun_etal-2023}. Given the manual DocRED training has been revised to produce ReDocRED, UGDRE generates two denoised distant datasets: one exploiting the DocRED manual as the training dataset (UGDRE-ReDocRED) and the other using the ReDocRED training dataset (UGDRE-ReDocRED). The second denoising strategy is DOREMI, an active learning system tailored for enhancing long-tail triples~\cite{menotti_etal-2026-DOREMI}. Like UGDRE, DOREMI exploits a set of \ac{DocRE} models pre-trained on a manual training dataset. Thus, this approach also generates two denoised versions of DocRED distant: DOREMI-DocRED and DOREMI-ReDocRED. 

\begin{table*}[t!]
\resizebox{\textwidth}{!}{%
    \centering
    \begin{tabular}{l|c|llll}
    \toprule
    Dataset & \multicolumn{1}{c|}{Instances} & \multicolumn{1}{c}{Invalid Triples ($\delta$)} & \multicolumn{1}{c}{Missing Inverse ($\sigma$)} & \multicolumn{1}{c}{Asymmetric Inc. ($\gamma$)} & \multicolumn{1}{c}{Cardinality Viol. ($\theta$)} \\ 
    \midrule
    UGDRE-DocRED & 1,717,161 & 51,671 (3.01\%) & 112,912 (53.50\%) & 13,026 (0.77\%) & 74 (0.03\%) \\
    DOREMI-DocRED & 1,704,161 & 29,799 (1.75\%) & 128,064 (75.13\%) & 6,030 (0.36\%) & 1,179 (0.73\%) \\
    UGDRE-ReDocRED & 3,379,590 & 94,290 (2.79\%) & 122,913 (25.42\%) & 19,416 (0.59\%) & 1,426 (0.52\%) \\
    DOREMI-ReDocRED & 3,957,238 & 69,035 (1.74\%) & 111,089 (22.00\%) & 15,244 (0.39\%) & 3,428 (1.43\%) \\
    \bottomrule
    \end{tabular}%
}
\caption{Structural consistency analysis of the denoised distant datasets. For each rule, we report the absolute number of errors in the dataset and its percentage.}
\label{tbl:den-data-quality}
\end{table*}
\subsection{Consistency Assessment}
% DocRED DS invalid 3.07% inverse 40.26% asymmetric 0.84% cardinality 0.01%
Table~\ref{tbl:den-data-quality} reports the structural consistency of the four distant datasets, produced by performing the same syntactic analysis described in Section~\ref{subsec:quality}. 

The percentage of invalid triples is almost doubled in UGDRE-based datasets than DOREMI. In particular, UGDRE-DocRED has almost the same percentage of invalid triples of the DocRED distant ($3.01\%$ vs $3.07\%$), hinting at the inability of current denoising techniques to eliminate structural inconsistencies. About missing inverse relations, the denoised datasets generated by using the DocRED training dataset misses much more inverse triples than the two datasets that uses ReDocRED. Indeed, the percentage of missing inverse relations in UGDRE-RedocRED is half of UGDRE-DocRED ($25.42\%$ vs $53.50\%$) while it is three times smaller between DOREMI-ReDocRED and DOREMI-DocRED ($25.42\%$ vs $75.13\%$). This is expected given that ReDocRED solves the false negative issues of DocRED~\cite{tan_etal-2022b}. 
Asymmetric inconsistencies are more evident in UGDRE datasets than the DOREMI ones. For instance, the percentage of asymmetric inconsistencies of UGDRE-DocRED is double than DOREMI-DocRED ($0.77\%$ vs $0.36\%$). Nevertheless, all denoised datasets reports a smaller percentage of asymmetric inconsistencies than DocRED distant (i.e., $0.84\%$). The percentage of cardinality violations are much larger than DocRED distant, which was almost zero ($0.01\%$). DOREMI datasets shows the highest number of errors, peaking at $1.43\%$ for DOREMI-ReDocRED. In general, for both denoising strategies, the DocRED-based denoised datasets have way less cardinality violations than the ReDocRED-based. In particular, UGDRE-DocRED has almost seventeen times less violations than UGDRE-ReDocRED ($0.03\%$ vs $0.52\%$), while DOREMI-DocRED has half violations than DOREMI-ReDocRED ($0.73\%$ vs $1.43\%$). This behavior may be attributed to the different dataset construction technique with respect to DocRED distant and to the fact that ReDocRED training dataset has much more training triples than the DocRED manual training dataset ($85,932$ vs $38,180$), training the model to predict more positive triples.

Overall, state-of-the-art denoising techniques fail to mitigate structural inconsistencies and, in the case of cardinality violations, even amplify them.

\begin{table}[t!]
\resizebox{\columnwidth}{!}{%
    \centering
    \begin{tabular}{c|l|llll}
    \toprule
    Model & Training Data & \multicolumn{1}{c}{Invalid ($\delta$)} & \multicolumn{1}{c}{Inverse ($\sigma$)} & \multicolumn{1}{c}{Asymmetric ($\gamma$)} \\
    \midrule
    \multirow{8}{*}{\begin{tabular}[c]{@{}l@{}}ATLOP \\ BERT\end{tabular}} & UGDRE-Doc & 106 (1.36\%) & 528 (59.26\%) & 38 (0.49\%) \\
    & \cellcolor{ours} Corrected & \cellcolor{ours} \textbf{42 (0.54\%)} & \cellcolor{ours} \textbf{65 (5.40\%)} & \cellcolor{ours} \textbf{2 (0.03\%)} \\
    \cmidrule{2-5}
    & DOREMI-Doc & 47 (0.72\%) & 335 (79.01\%) & 26 (0.40\%) \\
    & \cellcolor{ours} Corrected & \cellcolor{ours} \textbf{33 (0.46\%)} & \cellcolor{ours} \textbf{90 (10.79\%)} & \cellcolor{ours} \textbf{10 (0.14\%)} \\
    \cmidrule{2-5}
    & UGDRE-ReD & 188 (1.24\%) & 579 (26.82\%) & 66 (0.44\%) \\
    & \cellcolor{ours} Corrected & \cellcolor{ours} \textbf{118 (0.79\%)} & \cellcolor{ours} \textbf{149 (6.16\%)} & \cellcolor{ours} \textbf{6 (0.04\%)} \\
    \cmidrule{2-5}
    & DOREMI-ReD & 164 (1.02\%) & 381 (19.44\%) & 50 (0.31\%) \\
     & \cellcolor{ours} Corrected & \cellcolor{ours} \textbf{144 (0.96\%)} & \cellcolor{ours} \textbf{158 (8.27\%)} & \cellcolor{ours} \textbf{14 (0.09\%)} \\
    %\specialrule{1pt}{0pt}{2pt}
    \midrule
    \multirow{8}{*}{\begin{tabular}[c]{@{}l@{}}ATLOP \\ RoBERTa\end{tabular}} & UGDRE-Doc & 110 (1.41\%) & 505 (56.11\%) & 42 (0.54\%) \\
    & \cellcolor{ours} Corrected & \cellcolor{ours} \textbf{39 (0.48\%)} & \cellcolor{ours} \textbf{65 (5.15\%)} & \cellcolor{ours} \textbf{0 (0.00\%)} \\
    \cmidrule{2-5}
    & DOREMI-Doc & 48 (0.74\%) & 365 (74.95\%) & 16 (0.25\%) \\
    & \cellcolor{ours} Corrected & \cellcolor{ours} \textbf{37 (0.53\%)} & \cellcolor{ours} \textbf{93 (12.35\%)} & \cellcolor{ours} \textbf{10 (0.14\%)} \\
    \cmidrule{2-5}
    & UGDRE-ReD & 197 (1.29\%) & 621 (28.19\%) & 62 (0.41\%) \\
    & \cellcolor{ours} Corrected & \cellcolor{ours}\textbf{ 99 (0.64\%)} & \cellcolor{ours} \textbf{121 (4.73\%)} & \cellcolor{ours} \textbf{4 (0.03\%)} \\
    \cmidrule{2-5}
    & DOREMI-ReD & 194 (1.19\%) & 407 (20.65\%) & 60 (0.37\%) \\
    & \cellcolor{ours} Corrected & \cellcolor{ours} \textbf{120 (0.79\%)} & \cellcolor{ours} \textbf{181 (9.29\%)} & \cellcolor{ours} \textbf{12 (0.08\%)} \\
    \specialrule{1pt}{0pt}{2pt}
    \multirow{8}{*}{\begin{tabular}[c]{@{}l@{}}DREEAM \\ BERT\end{tabular}} & UGDRE-Doc & 106 (1.36\%) & 528 (59.26\%) & 38 (0.49\%) \\
    & \cellcolor{ours} Corrected & \cellcolor{ours} \textbf{27 (0.39\%)} & \cellcolor{ours} \textbf{71 (7.63\%)} & \cellcolor{ours} \textbf{4 (0.06\%)} \\
    \cmidrule{2-5}
    & DOREMI-Doc & 69 (1.04\%) & 363 (69.67\%) & 26 (0.40\%) \\
    & \cellcolor{ours} Corrected & \cellcolor{ours} \textbf{32 (0.46\%)} & \cellcolor{ours} \textbf{58 (7.81\%)} & \cellcolor{ours} \textbf{10 (0.14\%)} \\
    \cmidrule{2-5}
    & UGDRE-ReD & 188 (1.24\%) & 579 (26.82\%) & 66 (0.44\%) \\
    & \cellcolor{ours} Corrected & \cellcolor{ours} \textbf{87 (0.59\%)} & \cellcolor{ours} \textbf{158 (6.55\%)} & \cellcolor{ours} \textbf{12 (0.08\%)} \\
    \cmidrule{2-5}
    & DOREMI-ReD & 135 (0.85\%) & 287 (15.41\%) & 42 (0.27\%) \\
    & \cellcolor{ours} Corrected & \cellcolor{ours} \textbf{119 (0.73\%)} & \cellcolor{ours} \textbf{142 (6.59\%)} & \cellcolor{ours} \textbf{12 (0.07\%)} \\
    \midrule
    \multirow{8}{*}{\begin{tabular}[c]{@{}l@{}}DREEAM \\ RoBERTa\end{tabular}} & UGDRE-Doc & 115 (1.37\%) & 576 (56.92\%) & 110 (1.32\%) \\
    & \cellcolor{ours} Corrected & \cellcolor{ours} \textbf{27 (0.36\%)} & \cellcolor{ours} \textbf{86 (7.96\%)} & \cellcolor{ours} \textbf{10 (0.13\%)} \\
    \cmidrule{2-5}
    & DOREMI-Doc & 54 (0.71\%) & 461 (76.71\%) & 36 (0.48\%) \\
    & \cellcolor{ours} Corrected & \cellcolor{ours} \textbf{52 (0.68\%)} & \cellcolor{ours} \textbf{111 (9.85\%)} & \cellcolor{ours} \textbf{24 (0.32\%)} \\
    \cmidrule{2-5}
    & UGDRE-ReD & 116 (0.78\%) & 451 (22.09\%) & 76 (0.51\%) \\
    & \cellcolor{ours} Corrected & \cellcolor{ours} \textbf{73 (0.48\%)} & \cellcolor{ours} \textbf{129 (5.16\%)} & \cellcolor{ours} \textbf{30 (0.20\%)} \\
    \cmidrule{2-5}
    & DOREMI-ReD & 103 (0.64\%) & 306 (16.59\%) & 48 (0.30\%) \\
    & \cellcolor{ours} Corrected & \cellcolor{ours} \textbf{100 (0.61\%)} & \cellcolor{ours} \textbf{118 (5.45\%)} & \cellcolor{ours} \textbf{18 (0.11\%)} \\
    \bottomrule
    \end{tabular}%
}
\caption{Structural consistency of DocRE models predictions. The trained on different denoised distant datasets and evaluated on the \textbf{ReDocRED test dataset}. For space reasons, we use suffix ``-Doc'' as a diminutive for DocRED and ``-ReD'' as a diminutive for ReDocRED. For each rule, we report the absolute number of errors in the predictions and its percentage. Column ``\emph{Invalid}'' refers to invalid triples, ``\emph{Inverse}'' to missing inverse triples, and ``\emph{Asymmetric}'' to asymmetric inconsistencies.}
\label{tbl:den-train-preds}
\end{table}
\subsection{Training effect}
This section describes the structural consistency of the predictions of ATLOP and DREEAM trained on the four denoised datasets and their \emph{corrected} versions and their performance. 
The correction of the datasets follows the procedure described in Section~\ref{sub:cleaning} and the adopted experimental setup is described in Section~\ref{sub:setup}.

\subsubsection{Models Predictions}
Table~\ref{tbl:den-train-preds} reports the structural consistency of the predictions of ATLOP and DREEAM trained on the four denoised datasets and their \emph{corrected} versions. As for DocRED, training without structural noise systematically lowers the presence of ill-formatted triples in the predictions.

The percentage of missing inverse triples exhibits the most evident improvements, with corrected DOREMI-DocRED managing to reduce missing inverse relations by $-86.45\%$ on average in both \ac{DocRE} models and configurations. The improvements in ReDocRED-based datasets for missing inverse triples are more contained since the percentage of errors in the datasets was already much lower than DocRED-based datasets, but still ranging from $-55.01\%$ to $-83.22\%$ percentage points. 

The improvements in invalid triples and asymmetric inconsistencies are similar to those reported for DocRED. The highest improvement for invalid triples is in UGDRE-DocRED for DREEAM-RoBERTa, where the percentage of errors drops from $1.37\%$ to $0.36\%$, registering a relative reduction of $-73.72\%$. The same configuration and dataset shows the biggest reduction in asymmetric inconsistencies, going from $1.32\%$ to $0.13\%$ ($-90.15\%$ relative decrease). In addition, training ATLOP-RoBERTa with the corrected version of UGDRE-DocRED completely eliminates the presence of asymmetric inconsistencies in the predictions.

\begin{table*}[t!]
\resizebox{\textwidth}{!}{%
    \centering
    \begin{tabular}{c|l|lllll}
    \toprule
    Model & Training Data & \multicolumn{1}{c}{Precision} & \multicolumn{1}{c}{IgnPrec} & \multicolumn{1}{c}{Recall} & \multicolumn{1}{c}{F1} & \multicolumn{1}{c}{IgnF1} \\
    \midrule
    \multirow{8}{*}{\begin{tabular}[c]{@{}l@{}}ATLOP \\ BERT\end{tabular}} & UGDRE-Doc & 0.7881 & 0.6592 & 0.3512 & 0.4858 & 0.4582 \\
    & \cellcolor{ours} Corrected & \cellcolor{ours} \textbf{0.8002} \inp{1.54} & \cellcolor{ours} \textbf{0.6791} \inp{3.02} &  \cellcolor{ours} \textbf{0.3593} \inp{2.31} & \cellcolor{ours} \textbf{0.4959} \inp{2.08} & \cellcolor{ours} \textbf{0.4700} \inp{2.58} \\
    \cmidrule{2-7}
    & DOREMI-Doc & \textbf{0.8686} & \textbf{0.7903} & 0.3234 & 0.4702 & 0.4589 \\
    & \cellcolor{ours} Corrected & \cellcolor{ours} 0.8439 \dep{2.84} & \cellcolor{ours} 0.7716 \dep{2.37} & \cellcolor{ours} \textbf{0.3458} \inp{6.93} & \cellcolor{ours} \textbf{0.4905} \inp{4.32} & \cellcolor{ours} \textbf{0.4776} \inp{4.07} \\
    \cmidrule{2-7}
    & UGDRE-ReD & 0.8044 & \textbf{0.7445} & \textbf{0.7007} & \textbf{0.7490} & \textbf{0.7220}\\
    & \cellcolor{ours} Corrected & \cellcolor{ours} \textbf{0.8069} \inp{0.31} & \cellcolor{ours} 0.7093 \dep{4.73} & \cellcolor{ours} 0.6937 \dep{1.00} & \cellcolor{ours} 0.7461 \dep{0.39} & \cellcolor{ours} 0.7014 \dep{2.85} \\
    \cmidrule{2-7}
    & DOREMI-ReD & 0.7480 & 0.6297 & \textbf{0.6914} & \textbf{0.7186} & 0.6591 \\
    & \cellcolor{ours} Corrected & \cellcolor{ours} \textbf{0.7702} \inp{2.97} & \cellcolor{ours} \textbf{0.6650} \inp{5.61} & \cellcolor{ours} 0.6602 \dep{4.51} & \cellcolor{ours} 0.7110 \dep{1.06} & \cellcolor{ours} \textbf{0.6626} \inp{0.53} \\
    \midrule
    \multirow{8}{*}{\begin{tabular}[c]{@{}l@{}}ATLOP \\ RoBERTa\end{tabular}} & UGDRE-Doc & 0.7977 & 0.6761 & 0.3573 & 0.4936 & 0.4676 \\
    & \cellcolor{ours} Corrected & \cellcolor{ours} \textbf{0.8012} \inp{0.44} & \cellcolor{ours} \textbf{0.6832} \inp{1.05} & \cellcolor{ours} \textbf{0.3755} \inp{5.09} & \cellcolor{ours} \textbf{0.5114} \inp{3.61} & \cellcolor{ours} \textbf{0.4847} \inp{3.66} \\
    \cmidrule{2-7}
    & DOREMI-Doc & \textbf{0.8701} & \textbf{0.8031} & 0.3228 & 0.4709 & 0.4605 \\
    & \cellcolor{ours} Corrected & \cellcolor{ours} 0.8605 \dep{1.10} & \cellcolor{ours} 0.7942 \dep{1.11} & \cellcolor{ours} \textbf{0.3439} \inp{6.54} & \cellcolor{ours} \textbf{0.4914} \inp{4.35} & \cellcolor{ours} \textbf{0.4800} \inp{4.23} \\
    \cmidrule{2-7}
    & UGDRE-ReD & \textbf{0.8090} & \textbf{0.7502} & 0.7087 & \textbf{0.7555} & \textbf{0.7288} \\
    & \cellcolor{ours} Corrected & \cellcolor{ours} 0.8025 \dep{0.80} & \cellcolor{ours} 0.7048 \dep{6.05} & \cellcolor{ours} \textbf{0.7129} \inp{0.59} & \cellcolor{ours} 0.7551 \dep{0.05} & \cellcolor{ours} 0.7088 \dep{2.74} \\
    \cmidrule{2-7}
    & DOREMI-ReD & 0.7486 & 0.6311 & \textbf{0.7023} & \textbf{0.7247} & 0.6648 \\
    & \cellcolor{ours} Corrected & \cellcolor{ours} \textbf{0.7665} \inp{2.39} & \cellcolor{ours} \textbf{0.6624} \inp{4.96} & \cellcolor{ours} 0.6704 \dep{4.54} & \cellcolor{ours} 0.7152 \dep{1.31} & \cellcolor{ours} \textbf{0.6664} \inp{0.24} \\
    \specialrule{1pt}{0pt}{2pt}
    \multirow{8}{*}{\begin{tabular}[c]{@{}l@{}}DREEAM \\ BERT\end{tabular}} & UGDRE-Doc & 0.8483 & 0.7507 & 0.3247 & 0.4696 & 0.4533 \\
    & \cellcolor{ours} Corrected & \cellcolor{ours} \textbf{0.8562} \inp{0.93} & \cellcolor{ours} \textbf{0.7680} \inp{2.30} & \cellcolor{ours} \textbf{0.3413} \inp{5.12} & \cellcolor{ours} \textbf{0.4881} \inp{3.93} & \cellcolor{ours} \textbf{0.4726} \inp{4.25} \\
    \cmidrule{2-7}
    & DOREMI-Doc & 0.8796 & 0.8176 & 0.3334 & 0.4836 & 0.4737 \\
    & \cellcolor{ours} Corrected & \cellcolor{ours} \textbf{0.8853} \inp{0.64} & \cellcolor{ours} \textbf{0.8283} \inp{1.31} & \cellcolor{ours} \textbf{0.3564} \inp{6.88} & \cellcolor{ours} \textbf{0.5082} \inp{5.08} & \cellcolor{ours} \textbf{0.4983} \inp{5.20} \\
    \cmidrule{2-7}
    & UGDRE-ReD & \textbf{0.8265} & \textbf{0.7740} & \textbf{0.6966} & \textbf{0.7560} & \textbf{0.7333} \\
    & \cellcolor{ours} Corrected & \cellcolor{ours} 0.8180 \dep{1.03} & \cellcolor{ours} 0.7281 \dep{5.04} & \cellcolor{ours} 0.6958 \dep{0.12} & \cellcolor{ours} 0.7520 \dep{0.54} & \cellcolor{ours} 0.7116 \dep{2.96} \\
    \cmidrule{2-7}
    & DOREMI-ReD & \textbf{0.7701} & \textbf{0.6576} & 0.7037 & \textbf{0.7354} & 0.6799 \\
    & \cellcolor{ours} Corrected & \cellcolor{ours} 0.7617 \dep{1.09} & \cellcolor{ours} 0.6574 \dep{0.04} & \cellcolor{ours} \textbf{0.7083} \inp{0.65} & \cellcolor{ours} 0.7341 \dep{0.19} & \cellcolor{ours} \textbf{0.6819} \inp{0.29} \\
    \midrule
    \multirow{8}{*}{\begin{tabular}[c]{@{}l@{}}DREEAM \\ RoBERTa\end{tabular}} & UGDRE-Doc & 0.7806 & 0.6749 & \textbf{0.3748} & 0.5065 & 0.4820 \\
    & \cellcolor{ours} Corrected & \cellcolor{ours} \textbf{0.8575} \inp{9.85} & \cellcolor{ours} \textbf{0.7745} \inp{14.75} & \cellcolor{ours} 0.3692 \dep{1.51} & \cellcolor{ours} \textbf{0.5161} \inp{1.91} & \cellcolor{ours} \textbf{0.5000} \inp{3.74} \\
    \cmidrule{2-7}
    & DOREMI-Doc & 0.8737 & 0.8131 & 0.3782 & 0.5279 & 0.5163 \\
    & \cellcolor{ours} Corrected & \cellcolor{ours} \textbf{0.8815} \inp{0.89} & \cellcolor{ours} \textbf{0.8249} \inp{1.45} & \cellcolor{ours} \textbf{0.3876} \inp{2.47} & \cellcolor{ours} \textbf{0.5384} \inp{1.99} & \cellcolor{ours} \textbf{0.5273} \inp{2.14} \\
    \cmidrule{2-7}
    & UGDRE-ReD & \textbf{0.8464} & \textbf{0.7996} & 0.7252 & \textbf{0.7811} & 0.7606 \\
    & \cellcolor{ours} Corrected & \cellcolor{ours} 0.8329 \dep{1.58} & \cellcolor{ours} 0.7500 \dep{6.20} & \cellcolor{ours} \textbf{0.7321} \inp{0.96} & \cellcolor{ours} 0.7793 \dep{0.23} & \cellcolor{ours} \textbf{0.7410} \dep{2.58} \\
    \cmidrule{2-7}
    & DOREMI-ReD & \textbf{0.7939} & \textbf{0.6894} & 0.7267 & \textbf{0.7588} & \textbf{0.7075} \\
    & \cellcolor{ours} Corrected & \cellcolor{ours} 0.7785 \dep{1.95} & \cellcolor{ours} 0.6801 \dep{1.36} & \cellcolor{ours} \textbf{0.7334} \inp{0.91} & \cellcolor{ours} 0.7553 \dep{0.48} & \cellcolor{ours} 0.7057 \dep{0.27} \\
    \bottomrule
    \end{tabular}%
}
\caption{ATLOP and DREEAM trained on different denoised distant datasets and evaluated on the \textbf{ReDocRED test dataset}. For space reasons, we use suffix ``-Doc'' as a diminutive for DocRED and ``-ReD'' as a diminutive for ReDocRED.}
\label{tbl:den-train-eval}
\end{table*}
%\subsection{Training effect in performance}

\subsubsection{Models Performance}
Table~\ref{tbl:den-train-eval} reports the performance of ATLOP and DREEAM trained on the different denoised datasets and their corrected versions evaluated on the ReDocRED test dataset. While the correction of the DocRED distant supervision dataset produced systematic performance gains across both models and configurations, the denoised datasets did not always benefit from it. However, such a behavior may be attributed to the evaluation dataset containing some ill-formed triples (see Section~\ref{sec:data-quality}). Overall, there is a performance improvement in at least one metric for all models and configurations, except for DREEAM-BERT trained on the corrected version of UGDRE-ReDocRED. In 47 out of 80 cases ($59\%$), our approach yields performance improvements, demonstrating a net positive effect overall. Correcting the DocRED-based denoised datasets exhibit a performance improvement in more cases than ReDocRED-based datasets. Considering DocRED-based datasets, precision and ignored precision (\textsc{IgnPrec}) shows an improvement in six cases out of eight -- with a peak of $+9.85\%$ and $+14.75\%$ respectively for DREEAM-RoBERTa trained on the corrected UGDRE-DocRED -- recall in seven, while F1 and ignored F1 (\textsc{IgnF1}) increase for all models and configurations.

\begin{table*}[t!]
\resizebox{\textwidth}{!}{%
    \centering
    \begin{tabular}{c|c|l|lllll}
    \toprule
    & PLM & Training Data & Precision & IgnPrec & Recall & F1 & IgnF1 \\
    \midrule
    \multirow{10}{*}{\rotatebox[origin=c]{90}{ATLOP}} & \multirow{5}{*}{BERT} & DocRED DS & 0.7657 & 0.6069 & 0.2983 & 0.4293 & 0.4000 \\
    \cmidrule{3-8}
    & & w/o invalid & 0.7684 \inp{0.35} & 0.6196 \inp{2.08} & 0.3009 \inp{0.87} & 0.4324 \inp{0.73} & 0.4050 \inp{1.27} \\
    & & w/o inverse & 0.7605 \dep{0.68} & 0.5980 \dep{1.46} & 0.3088 \inp{3.55} & 0.4393 \inp{2.33} & 0.4073 \inp{1.84} \\
    & & w/o asymmetric & 0.7660 \inp{0.03} & 0.6094 \inp{0.41} & 0.3009 \inp{0.89} & 0.4321 \inp{0.65} & 0.4029 \inp{0.73} \\
    \cmidrule{3-8}
    & & \cellcolor{ours} All rules & \cellcolor{ours} \textbf{0.7704} \inp{0.60} & \cellcolor{ours} \textbf{0.6197} \inp{2.11} & \cellcolor{ours} \textbf{0.3103} \inp{4.05} & \cellcolor{ours} \textbf{0.4424} \inp{3.06} & \cellcolor{ours} \textbf{0.4136} \inp{3.41} \\
    \cmidrule{2-8}
    & \multirow{5}{*}{RoBERTa} & DocRED DS & 0.7603 & 0.6041 & 0.3106 & 0.4410 & 0.4102 \\
    \cmidrule{3-8}
    & & w/o invalid & 0.7623 \inp{0.26} & 0.6267 \inp{3.74} & 0.3184 \inp{2.53} & 0.4492 \inp{1.86} & 0.4223 \inp{2.94} \\
    & & w/o inverse & 0.7595 \dep{0.11} & 0.6061 \inp{0.32} & \textbf{0.3305} \inp{6.43} & \textbf{0.4606} \inp{4.44} & \textbf{0.4277} \inp{4.27} \\
    & & w/o asymmetric & 0.7648 \inp{0.59} & 0.6093 \inp{0.85} & 0.3105 \dep{0.04} & 0.4416 \inp{0.14} & 0.4113 \inp{0.26} \\
    \cmidrule{3-8}
    & & \cellcolor{ours} All rules & \cellcolor{ours} \textbf{0.7806} \inp{2.66} & \cellcolor{ours} \textbf{0.6390} \inp{5.77} & \cellcolor{ours} 0.3194 \inp{2.85} & \cellcolor{ours} 0.4533 \inp{2.80} & \cellcolor{ours} 0.4259 \inp{3.82} \\
    \specialrule{1pt}{0pt}{2pt}
    \multirow{10}{*}{\rotatebox[origin=c]{90}{DREEAM}} & \multirow{5}{*}{BERT} & DocRED DS & 0.8116 & \textbf{0.7667} & 0.2613 & 0.3953 & 0.3897 \\
    \cmidrule{3-8}
    & & w/o invalid & 0.7970 \dep{1.80} & 0.6847 \dep{10.69} & 0.2747 \inp{5.14} & 0.4086 \inp{3.36} & 0.3921 \inp{0.61} \\
    & & w/o inverse & 0.8109 \dep{0.09} & 0.6839 \dep{10.80} & 0.2772 \inp{6.09} & 0.4132 \inp{4.52} & 0.3945 \inp{1.22} \\
    & & w/o asymmetric & \textbf{0.8148} \inp{0.39} & 0.6888 \dep{10.16} & 0.2635 \inp{0.86} & 0.3983 \inp{0.75} & 0.3812 \dep{2.19} \\
    \cmidrule{3-8}
    & & \cellcolor{ours} All rules & \cellcolor{ours} 0.7992 \dep{1.54} & \cellcolor{ours} 0.6909 \dep{9.89} & \cellcolor{ours} \textbf{0.3018} \inp{15.52} & \cellcolor{ours} \textbf{0.4382} \inp{10.85} & \cellcolor{ours} \textbf{0.4201} \inp{7.80} \\
    \cmidrule{2-8}
    & \multirow{5}{*}{RoBERTa} & DocRED DS & 0.8270 & \textbf{0.7885} & 0.2864 & 0.4255 & 0.4202 \\
    \cmidrule{3-8}
    & & w/o invalid & 0.8220 \dep{0.60} & 0.7184 \dep{8.90} & 0.2950 \inp{2.99} & 0.4342 \inp{2.04} & 0.4183 \dep{0.47} \\
    & & w/o inverse & 0.8252 \dep{0.21} & 0.7102 \dep{9.93} & 0.2937 \inp{2.52} & 0.4332 \inp{1.81} & 0.4155 \dep{1.12} \\
    & & w/o asymmetric & \textbf{0.8295} \inp{0.31} & 0.7158 \dep{9.23} & 0.2787 \dep{2.71} & 0.4172 \dep{1.95} & 0.4012 \dep{4.53} \\
    \cmidrule{3-8}
    & & \cellcolor{ours} All rules & \cellcolor{ours} 0.8213 \dep{0.68} & \cellcolor{ours} 0.7179 \dep{8.96} & \cellcolor{ours} \textbf{0.3066} \inp{7.05} & \cellcolor{ours} \textbf{0.4466} \inp{4.95} & \cellcolor{ours} \textbf{0.4297} \inp{2.26}\\
    \bottomrule
    \end{tabular}%
}
\caption{Models trained on DocRED distant (``DocRED DS'') and its structurally-consistent versions and evaluated on the \textbf{ReDocRED dev dataset}. Row ``All rules" considers the dataset cleaned from all structural constraints, row ``w/o invalid" considers the dataset dataset cleaned from invalid triples only, row ``w/o inverse" dataset cleaned from missing inverse triples only, and row ``w/o asymmetric" dataset cleaned from asymmetric inconsistencies only. The percentage performance improvement compared to DocRED distant is reported between parentheses.}
\label{tbl:ablation}
\end{table*}

\section{Individual Structural Constraint Impact}
\label{app:ablation}
We performed a study on the impact of each noise source to decide which structural inconsistencies to include in the study. In particular, we trained ATLOP on different versions of the DocRED distant dataset, where we corrected a single structural inconsistency (invalid triples, missing inverse, or asymmetric inconsistency). Table~\ref{tbl:ablation} reports the micro-averaged performance of ATLOP and DREEAM trained on the different versions of the DocRED distant dataset evaluated on the development dataset. We report the performance of the original distant dataset (row ``DocRED DS"). We also report the performance when cleaning only a single constraints. For instance, row ``w/o invalid" reports the performance of ATLOP and DREEAM trained on the DocRED distant dataset cleaned from invalid triples only. %Row ``All rules" reports the performance of the models trained on the corrected dataset.

Correcting individual sources of noise yields consistent performance gains for ATLOP in both transformer configurations, with the magnitude of improvement varying by noise type. Invalid triples represent the predominant source of noise; for instance, in ATLOP-BERT, correcting invalid triples of the DocRED distant dataset results in a precision improvement of $+0.35\%$, recall $+0.87\%$, and F1 $+0.73\%$. The gain is even more pronounced in ignored metrics (\textsc{ignPrec} $+2.08\%$ and \textsc{ignF1} $+1.27\%$), when entity pairs seen during training are discarded. Addressing the missing inverse triples results in the highest recall increase ($+3.55\%$ in ATLOP-BERT), but shows a slight decrease in terms of precision ($-0.68\%$ in ATLOP-BERT and $-0.11\%$ in ATLOP-RoBERTa). Integrating all corrections cumulatively achieves the highest performance gains (precision $+0.60\%$, \textsc{ignPrec} $+2.11\%$, recall $+4.05\%$, F1 $+3.41\%$, \textsc{ignF1} $+3.41\%$ in ATLOP-BERT). For ATLOP-RoBERTa, correcting all structural inconsistencies registers the highest average improvement across all metrics. The configuration shows the best precision improvement (precision $+2.66\%$, \textsc{ignPrec} $+5.77\%$) and the second best gain in recall ($+2.85\%$), F1 ($+2.80\%$), and \textsc{ignF1} ($+3.82\%$).

Correcting different types of rules for DREEAM shows the same performance trend as in ATLOP; invalid rules are the most influential source of noise, and inverse rules boost the recall. 
Applying all structural constraints together achieves the best results, showing a significant performance improvement in recall ($+15.52\%$ in DREEAM-BERT, $+7.05\%$ for DREEAM-RoBERTa) and F1 ($+10.85\%$ in DREEAM-BERT, $+4.95\%$ for DREEAM-RoBERTa), while registering a small precision degradation ($-1.54\%$ in DREEAM-BERT, $-0.68\%$ for DREEAM-RoBERTa). 

Both \ac{DocRE} models and both configurations demonstrate that each source of noise has a significant impact on the model performance: removing invalid triples improves precision, while adding missing inverse relations increases recall. However, integrating all structural constraints together harnesses the strength of each individual rule, producing the best overall performance.

\section{DocRED Structural Constraints}
\label{app:rules}
This section reports the structural constraints exploited for our study. The rules are also available in JSON format in Github.~\footnote{Folder ``\texttt{data/rules/}" at: \url{https://github.com/mntlra/DocRE-StructuralConsistency}}
Tables~\ref{tab:domain-range-rules1} and~\ref{tab:domain-range-rules2} report the domain and range entity type allowed for each DocRED relation. Table~\ref{tab:inverses} reports the relations with an inverse in DocRED. To conclude, Table~\ref{tab:cardinality} reports the relations with an upper bound on the maximum cardinality.
\begin{table*}[t!]
    \centering
    \footnotesize
    \begin{tabularx}{\textwidth}{l|X|X|X}
    \toprule
    Wikidata ID & Name & Domain & Range \\
    \toprule
    P6 & head of government & LOC|ORG|MISC & PER|MISC \\ 
    P17 & country & ORG|LOC|PER|MISC|TIME & LOC|ORG|MISC \\ 
    P19 & place of birth & PER|MISC & LOC|MISC \\ 
    P20 & place of death & PER|MISC & LOC|MISC \\ 
    P22 & father & PER|MISC & PER|MISC \\ 
    P25 & mother & PER|MISC & PER|MISC \\ 
    P26 & spouse & PER|MISC & PER|MISC \\ 
    P27 & country of citizenship & PER|ORG|MISC & LOC|ORG|MISC \\ 
    P30 & continent & ORG|LOC|MISC & LOC|MISC \\ 
    P31 & instance of & PER|ORG|LOC|TIME|NUM|MISC & PER|ORG|LOC|TIME|NUM|MISC \\ 
    P35 & head of state & LOC|MISC & PER|MISC \\ 
    P36 & capital & LOC|ORG|MISC & LOC|MISC \\ 
    P37 & official language & LOC|ORG|MISC & MISC \\ 
    P39 & position held & PER|MISC & ORG|MISC \\ 
    P40 & child & PER|MISC & PER|MISC \\ 
    P50 & author & MISC & PER|ORG|MISC \\ 
    P54 & member of sports team & PER|NUM|MISC & ORG|LOC|MISC \\ 
    P57 & director & MISC & PER|MISC \\ 
    P58 & screenwriter & MISC & PER|MISC \\ 
    P69 & educated at & PER|MISC & ORG|LOC|MISC \\ 
    P86 & composer & NUM|MISC & PER|MISC \\ 
    P102 & member of political party & PER|MISC & ORG|MISC \\ 
    P108 & employer & PER|MISC & LOC|ORG|MISC \\ 
    P112 & founded by & LOC|ORG|MISC & PER|ORG|LOC|MISC \\ 
    P118 & league & PER|ORG|MISC & ORG|LOC|MISC \\ 
    P123 & publisher & MISC & PER|ORG|MISC \\ 
    P127 & owned by & ORG|LOC|PER|MISC & PER|ORG|LOC|MISC \\ 
    P131 & located in the administrative territorial entity & ORG|LOC|MISC & LOC|MISC \\ 
    P136 & genre & PER|ORG|MISC & ORG|LOC|MISC \\ 
    P137 & operator & LOC|ORG|MISC & ORG|LOC|PER|MISC \\ 
    P140 & religion & PER|ORG|LOC|MISC & ORG|MISC \\ 
    P150 & contains administrative territorial entity & ORG|LOC|MISC & ORG|LOC|MISC \\ 
    P155 & follows & PER|ORG|LOC|TIME|NUM|MISC & PER|ORG|LOC|TIME|NUM|MISC \\ 
    P156 & followed by & PER|ORG|LOC|TIME|NUM|MISC & PER|ORG|LOC|TIME|NUM|MISC \\ 
    P159 & headquarters location & ORG|LOC|MISC & ORG|LOC|MISC \\ 
    P161 & cast member & MISC & PER|MISC \\ 
    P162 & producer & MISC & PER|ORG|MISC \\ 
    P166 & award received & PER|LOC|ORG|MISC & MISC|ORG \\ 
    P170 & creator & LOC|ORG|PER|NUM|MISC & PER|ORG|MISC \\ 
    P171 & parent taxon & PER|ORG|LOC|TIME|NUM|MISC & PER|ORG|LOC|TIME|NUM|MISC \\ 
    P172 & ethnic group & PER|ORG|LOC|MISC & LOC|ORG|MISC \\ 
    P175 & performer & MISC|ORG|PER & PER|ORG|MISC \\ 
    P176 & manufacturer & MISC|ORG & ORG|PER|MISC \\ 
    P178 & developer & MISC & ORG|PER|MISC \\ 
    P179 & series & MISC|NUM & ORG|MISC \\ 
    P190 & sister city & LOC|MISC & LOC|MISC \\ 
    P194 & legislative body & LOC|ORG|MISC & LOC|ORG|MISC \\ 
    P205 & basin country & LOC|MISC & LOC|MISC \\
    \bottomrule
    \end{tabularx}
    \caption{Domain and Range constraints of DocRED relations (I). For each relation, we report the Wikidata ID and a textual description.}
    \label{tab:domain-range-rules1}
\end{table*}

\begin{table*}[t!]
    \centering
    \footnotesize
    \begin{tabularx}{\textwidth}{l|X|X|X}
    \toprule
    Wikidata ID & Name & Domain & Range \\
    \toprule
    P206 & located in or next to body of water & LOC|ORG|MISC & LOC|MISC \\ 
    P241 & military branch & PER|ORG|MISC & ORG|MISC \\ 
    P264 & record label & PER|ORG|MISC & ORG|MISC \\ 
    P272 & production company & ORG|MISC & ORG|MISC \\ 
    P276 & location & PER|ORG|LOC|TIME|NUM|MISC & ORG|LOC|MISC \\ 
    P279 & subclass of & PER|ORG|LOC|TIME|NUM|MISC & PER|ORG|LOC|TIME|NUM|MISC \\ 
    P355 & subsidiary & PER|ORG|MISC & LOC|ORG|MISC \\ 
    P361 & part of & PER|ORG|LOC|TIME|NUM|MISC & PER|ORG|LOC|TIME|NUM|MISC \\ 
    P364 & original language of work & MISC & LOC|MISC \\ 
    P400 & platform & MISC & MISC \\ 
    P403 & mouth of the watercourse & LOC|MISC & LOC|MISC \\ 
    P449 & original network & MISC & ORG|MISC \\ 
    P463 & member of & PER|ORG|LOC|MISC & LOC|ORG|MISC \\ 
    P488 & chairperson & ORG|LOC|MISC & PER|MISC \\ 
    P495 & country of origin & MISC & LOC|MISC \\ 
    P527 & has part & PER|ORG|LOC|TIME|NUM|MISC & PER|ORG|LOC|TIME|NUM|MISC \\ 
    P551 & residence & PER|MISC & LOC|MISC \\ 
    P569 & date of birth & PER|MISC & TIME|MISC \\ 
    P570 & date of death & PER|MISC & TIME|MISC \\ 
    P571 & inception & ORG|LOC|TIME|MISC & TIME|MISC \\ 
    P576 & dissolved, abolished or demolished & ORG|LOC|MISC & TIME|MISC \\ 
    P577 & publication date & MISC & TIME|MISC \\ 
    P580 & start time & PER|ORG|LOC|TIME|NUM|MISC & TIME|MISC \\ 
    P582 & end time & PER|ORG|LOC|TIME|NUM|MISC & TIME|MISC \\ 
    P585 & point in time & PER|ORG|LOC|TIME|NUM|MISC & TIME|MISC \\ 
    P607 & conflict & PER|ORG|LOC|MISC & ORG|TIME|LOC|MISC \\ 
    P674 & characters & MISC & PER|ORG|MISC|LOC \\ 
    P676 & lyrics by & MISC & PER|ORG|MISC \\ 
    P706 & located on terrain feature & LOC|ORG|MISC & LOC|MISC \\ 
    P710 & participant & PER|ORG|LOC|MISC & PER|ORG|LOC|MISC \\ 
    P737 & influenced by & PER|ORG|LOC|TIME|NUM|MISC & PER|ORG|LOC|TIME|NUM|MISC \\ 
    P740 & location of formation & ORG|MISC & LOC|MISC \\ 
    P749 & parent organization & LOC|ORG|MISC & PER|ORG|MISC \\ 
    P800 & notable work & PER|ORG|MISC & PER|LOC|ORG|NUM|MISC \\ 
    P807 & separated from & LOC|ORG|MISC & LOC|ORG|MISC \\ 
    P840 & narrative location & MISC & LOC|MISC \\ 
    P937 & work location & PER|ORG|MISC & LOC|MISC \\ 
    P1001 & applies to jurisdiction & PER|ORG|MISC & LOC|ORG|MISC \\ 
    P1056 & product or material produced & PER|ORG|MISC & MISC|ORG \\ 
    P1198 & unemployment rate & LOC|MISC & NUM|MISC \\ 
    P1336 & territory claimed by & LOC|MISC & PER|ORG|LOC|MISC \\ 
    P1344 & participant of & PER|ORG|LOC|MISC & PER|ORG|LOC|MISC \\ 
    P1365 & replaces & PER|ORG|LOC|MISC & PER|ORG|LOC|MISC \\ 
    P1366 & replaced by & PER|ORG|LOC|MISC & PER|ORG|LOC|MISC \\ 
    P1376 & capital of & LOC|MISC & LOC|ORG|MISC \\ 
    P1412 & languages spoken, written or signed & PER|MISC & LOC|MISC \\ 
    P1441 & present in work & PER|ORG|LOC|MISC & MISC \\ 
    P3373 & sibling & PER|MISC & PER|MISC \\
    \bottomrule
    \end{tabularx}
    \caption{Domain and Range constraints of DocRED relations (II). For each relation, we report the Wikidata ID and a textual description (column ``Name'').}
    \label{tab:domain-range-rules2}
\end{table*}

\begin{table*}[t!]
    \centering
    %\footnotesize
    \begin{tabularx}{\textwidth}{l|X|l|X}
    \toprule
    Wikidata ID & Name & Inverse ID & Inverse Name \\
    \toprule
    P22 & father & P40 & child \\ 
    P25 & mother & P40 & child \\ 
    P26 & spouse & P26 & spouse \\ 
    P36 & capital & P1376 & capital of \\ 
    P50 & author & P800 & notable work \\ 
    P57 & director & P800 & notable work \\ 
    P86 & composer & P800 & notable work \\ 
    P155 & follows & P156 & followed by \\ 
    P156 & followed by & P155 & follows \\ 
    P170 & creator & P800 & notable work \\ 
    P176 & manufacturer & P1056 & product or material produced \\ 
    P355 & subsidiary & P749 & parent organization \\ 
    P361 & part of & P527 & has part \\ 
    P527 & has part & P361 & part of \\ 
    P674 & characters & P1441 & present in work \\ 
    P710 & participant & P1344 & participant of \\ 
    P749 & parent organization & P355 & subsidiary \\ 
    P1344 & participant of & P710 & participant \\ 
    P1365 & replaces & P1366 & replaced by \\ 
    P1366 & replaced by & P1365 & replaces \\ 
    P1376 & capital of & P36 & capital \\ 
    P1441 & present in work & P674 & characters \\ 
    P3373 & sibling & P3373 & sibling \\ 
    \bottomrule
    \end{tabularx}
    \caption{DocRED relations with an inverse relation. For each relation, we report its Wikidata ID, its textual description (``Name'') and the Wikidata ID (``Inverse ID'') and description (``Inverse Name'') of its inverse relation. Note that inverse relations also include symmetric relations.}
    \label{tab:inverses}
\end{table*}

\begin{table*}[t!]
    \centering
    %\footnotesize
    \begin{tabular}{l|l|l}
    \toprule
    Wikidata ID & Name & Max Cardinality \\
    \toprule
    P19 & place of birth & 2 \\
    P20 & place of death & 2 \\
    P22 & father & 2 \\
    P25 & mother & 2 \\
    P569 & date of birth & 2 \\
    P570 & date of death & 2 \\
    P571 & inception & 2 \\
    P576 & dissolved, abolished or demolished & 2 \\
    P580 & start time & 2 \\
    P582 & end time & 2 \\
    P585 & point in time & 2 \\
    \bottomrule
    \end{tabular}
    \caption{DocRED relations with bounded maximum cardinality. For each relation, we report its Wikidata ID, its textual description (``Name'') and the Wikidata ID (``Inverse ID'') and description (``Inverse Name'') of its inverse relation. Note that inverse relations also include symmetric relations (Wikidata ID = Inverse ID).}
    \label{tab:cardinality}
\end{table*}

\newpage

\end{document}